\documentclass{bmvc2k}

\title{LAFR: Efficient Diffusion-based Blind Face Restoration via Latent Codebook Alignment Adapter}

\addauthor{Runyi Li}{lirunyi@stu.pku.edu.cn}{1}
\addauthor{Bin Chen}{chenbin@stu.pku.edu.cn}{1}
\addauthor{Jian Zhang}{zhangjian.sz@pku.edu.cn}{1}
\addauthor{Radu Timofte}{radu.timofte@uni-wuerzburg.de}{2}

\addinstitution{
 School of Electronic and\\
 Computer Engineering,\\
 Peking University,\\
 Shenzhen, China
}
\addinstitution{
 Computer Vision Lab,\\
 CAIDAS and IFI,\\
 University of W\"urzburg,\\
 W\"urzburg, Germany
}

\runninghead{Li et al.}{LAFR: Blind Face Restoration via Latent Codebook Alignment}

\def\eg{\emph{e.g}\bmvaOneDot}
\def\Eg{\emph{E.g}\bmvaOneDot}
\def\etal{\emph{et al}\bmvaOneDot}
\def\ie{\emph{i.e}\bmvaOneDot}
\def\etc{\emph{etc}\bmvaOneDot}
\def\cf{\emph{cf}\bmvaOneDot}

\usepackage{graphicx}
\usepackage{booktabs}
\usepackage{amsmath}
\usepackage{amssymb}
\usepackage{multicol}
\usepackage{multirow}
\usepackage{tabularx}
\usepackage{arydshln}
\usepackage{xcolor}
\usepackage{pifont}
\usepackage{placeins}
\usepackage{float}
\usepackage{wrapfig}
\usepackage{graphicx}
\usepackage{booktabs}
\usepackage{amsmath}
\usepackage{amssymb}
\usepackage{multirow}
\usepackage{tabularx}
\usepackage{arydshln}
\usepackage{xcolor}
\usepackage{pifont}
\usepackage{placeins}
\usepackage{float}
\usepackage{listings}
\usepackage[ruled,linesnumbered]{algorithm2e}
\SetKwComment{Comment}{/* }{ */}
\hypersetup{hypertexnames=false}

\hypersetup{
  pdftitle  = {LAFR: Efficient Diffusion-based Blind Face Restoration via Latent Codebook Alignment Adapter},
  pdfauthor = {Runyi Li, Bin Chen, Jian Zhang, Radu Timofte},
  pdfsubject = {BMVC 2026},
  pdfkeywords = {blind face restoration, latent diffusion,
                 codebook alignment, parameter-efficient fine-tuning}
}

\definecolor{Red}{RGB}{217,57,50}
\definecolor{UNetOrange}{RGB}{248,205,172}
\definecolor{UNetGray}{RGB}{166,166,166}
\newcommand{\best}[1]{\textcolor{Red}{\textbf{#1}}}

\newcommand{\sbest}[1]{\textcolor{blue}{\underline{#1}}}

\newcommand{\pnum}[1]{{\color{gray!60!black}\itshape #1}}

\definecolor{Red}{RGB}{217,57,50}

\definecolor{codegreen}{RGB}{0,128,0}
\definecolor{codegray}{RGB}{105,105,105}
\definecolor{codepurple}{RGB}{128,0,128}
\definecolor{backcolour}{RGB}{245,245,245}
\lstdefinestyle{pythonstyle}{
    backgroundcolor=\color{backcolour},
    commentstyle=\color{codegreen},
    keywordstyle=\color{codepurple},
    stringstyle=\color{codegreen},
    basicstyle=\ttfamily\scriptsize,
    breakatwhitespace=false,
    breaklines=true,
    captionpos=b,
    keepspaces=true,
    numbers=left,
    numbersep=5pt,
    showspaces=false,
    showstringspaces=false,
    showtabs=false,
    tabsize=2
}

\let\suppmaketitle\maketitle

\begin{document}

\maketitle

\begin{abstract}
Blind face restoration from low-quality (LQ) images is a challenging task that requires not only high-fidelity image reconstruction, but also preservation of facial identity. Although diffusion models like Stable Diffusion have shown promise in generating high-quality (HQ) images, their VAE modules are typically trained on broad natural-image data dominated by HQ content; severely degraded LQ inputs therefore yield latents that fall in low-density regions of the diffusion prior, weakening the effectiveness of LQ conditions during the denoising process. Existing approaches often tackle this issue by retraining the VAE encoder, which is computationally expensive and memory intensive.
To address this limitation efficiently, we propose \textbf{LAFR} (\textbf{L}atent \textbf{A}lignment for \textbf{F}ace \textbf{R}estoration), a codebook-based latent space adapter that aligns LQ latents to the in-distribution region the diffusion prior was trained to denoise, enabling sampling from a better matched latent condition without altering the original VAE. To further improve identity and structural consistency relative to OSEDiff-style baselines, we introduce a multilevel restoration loss combining constraints from semantic identity embeddings and facial structural priors. Furthermore, by leveraging the inherent structural regularity of facial images, we show that lightweight fine-tuning of a diffusion prior on just \textbf{0.9\%} of FFHQ can achieve competitive results while reducing training time by \textbf{70\%}.
We also give a short analysis of why a codebook correction closes a \emph{mode-dependent} latent gap that no global normalisation or affine map can. In summary, LAFR is a data- and compute-efficient one-step face-restoration method that improves perceptual quality, FID, landmark accuracy, and identity/structure consistency over OSEDiff-style baselines, without claiming uniform state of the art; it reaches this operating point while remaining competitive with far more expensive pipelines.
\end{abstract}

\section{Introduction}
Face image restoration is a critical task within the broader domain of image restoration~\cite{hu2020face, zhao2022rethinking, yang2021gan, chen2021progressive, varanka2024pfstorer}, with applications ranging from photo enhancement to facial forensics~\cite{9744329,menon2020pulse,yu2024scaling}. Unlike general images, facial images exhibit highly structured patterns and identity-sensitive features~\cite{varanka2024pfstorer,hu2021face,10888736}, necessitating tailored solutions to ensure structural fidelity and identity preservation.
Recent advances in diffusion models~\cite{ho2020denoising, song2020score, song2021denoising, Rombach_Blattmann_Lorenz_Esser_Ommer_2022} have positioned them as strong generative priors across various restoration tasks. Among these, latent diffusion models have gained popularity for face image reconstruction, where inputs are encoded into a latent space before restoration. The VAE encoder in Stable Diffusion~\cite{Rombach_Blattmann_Lorenz_Esser_Ommer_2022} is trained on broad natural-image data that is dominated by visually clean content; consequently, when severely degraded LQ inputs are encoded, the resulting latents drift into regions that are sparsely populated by the diffusion denoising prior, which itself is centred on the in-distribution latents seen during training. This provides poor starting conditions for the diffusion trajectory~\cite{wang2025osdface, LIU2024110659}. We term this the \textit{latent distribution mismatch} problem, and we quantify it via latent-space MMD/$\mu$-$\Sigma$ distance in Sec.~\ref{sec:latent_validation}, complementing the qualitative t-SNE evidence.

Existing approaches~\cite{chen2024faithdiff, wang2025osdface, suin2024clr} attempt to correct this by retraining the VAE and incorporating complex alignment modules, but at the cost of increased computational load and inference latency. Another direction fine-tunes the diffusion UNet on large face datasets (\eg FFHQ~\cite{karras2019style} with 70K images), demanding significant compute and memory resources.

These challenges motivate two questions: given the latent distribution mismatch, how can the diffusion process be guided effectively \emph{without} modifying the VAE; and can a pretrained diffusion model be adapted for face restoration by updating only a minimal subset of parameters on a small training set? We answer both with a solution requiring 600 training images (0.9\% of FFHQ) and 7.5M trainable parameters, bridging general diffusion priors and the demands of facial restoration with a small, explicitly measured adaptation module.

\begin{figure*}[t]
    \centering
    \includegraphics[width=0.80\linewidth]{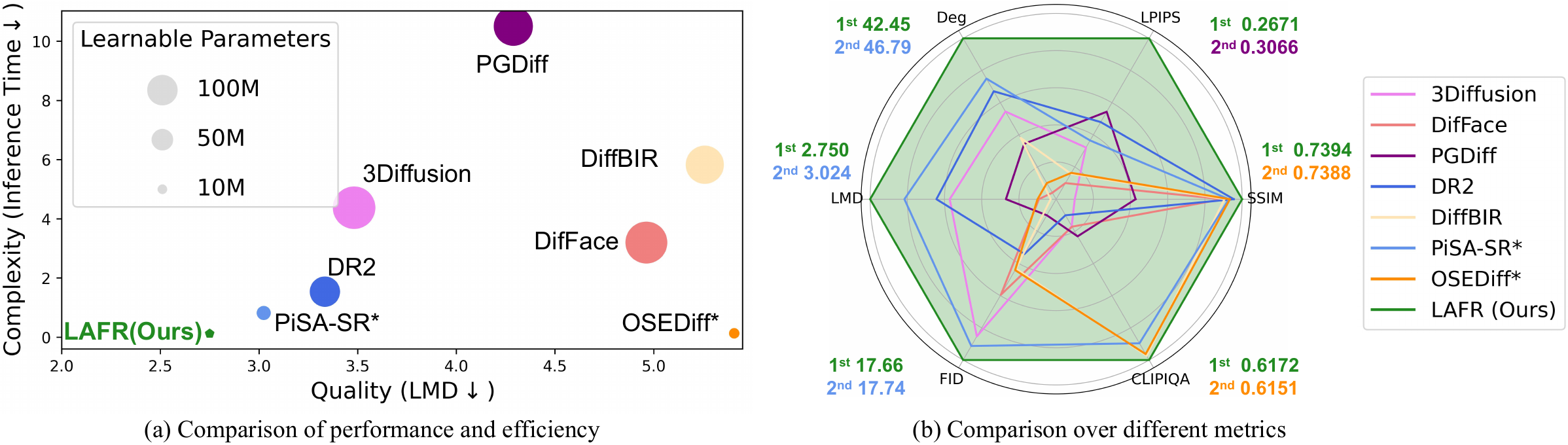}
\caption{Comparison of LAFR with representative face restoration methods.
(a) Performance and efficiency across different methods. Bubble size reflects learnable parameters. LAFR provides a favorable balance between restoration quality and computational cost.
(b) Quantitative comparison across multiple metrics. All metrics are normalized; for metrics where lower is better, we take the reciprocal for consistent visual interpretation. * means re-trained on FFHQ~\cite{karras2019style}.}
    \label{fig:teaser}
\end{figure*}

To address the latent distribution mismatch, we introduce a lightweight codebook-based alignment adapter~\cite{vqvae, Preechakul_2022_CVPR,liang2025diffusion} that corrects LQ latent codes toward the HQ distribution prior to denoising, leaving the VAE encoder and decoder intact, and, to improve identity preservation, a multilevel restoration loss enforcing consistency across appearance, semantic identity embeddings, and structural details. As shown in Fig.~\ref{fig:teaser}, LAFR offers a favorable balance between quality and efficiency. Our contributions are:

\noindent \ding{113}~(1) An efficient strategy that adapts a pretrained diffusion prior to face restoration using 600 training images and 7.5M trainable parameters, cutting training time by 70\% in our reproduced setting.

\noindent \ding{113}~(2) A lightweight codebook-based latent alignment adapter that corrects the LQ-to-HQ latent distribution mismatch without retraining the VAE, orthogonal to UNet fine-tuning, together with an analysis (Proposition~1) of \emph{why} a $K$-piece codebook correction closes a mode-dependent latent gap that no global normalisation or affine map can.

\noindent \ding{113}~(3) A multilevel restoration loss combining appearance, semantic identity embeddings, and structural alignment, improving identity and structural consistency over image-only losses against an OSEDiff-style one-step baseline.

\noindent \ding{113}~(4) Across synthetic and real-world benchmarks, LAFR improves perceptual quality, FID, landmark accuracy, and identity/structure consistency relative to OSEDiff-style one-step baselines, while greatly reducing training data and time; we do not claim uniform state of the art (\eg DiffBIR leads PSNR, GFPGAN leads ArcFace similarity).

\section{Related Work}

\noindent\textbf{Blind Face Restoration.}
Early blind face restoration methods leverage GAN-based priors to generate perceptually realistic details. GFPGAN~\cite{wang2021gfpgan} incorporates facial component dictionaries and spatial feature transform layers within a GAN framework. RestoreFormer++~\cite{wang2023restoreformerpp} uses multi-head cross-attention to query a learned HQ feature codebook. VQFR~\cite{gu2022vqfr} employs a VQ codebook to replace corrupted facial tokens with high-fidelity counterparts. Despite strong perceptual quality, GAN-based methods often suffer from mode collapse and identity drift. Diffusion-based methods~\cite{lu20243d, yang2023pgdiff, wang2023dr2, yue2024difface, diffbir, wang2023gan} subsequently achieve higher fidelity by leveraging large pretrained priors, at the cost of many denoising steps. One-step distillation approaches~\cite{wu2024osediff, sun2024pixel, interLCM2025} reduce inference cost while maintaining competitive quality, offering a more practical balance. Our method builds on this paradigm with additional focus on training efficiency.

\noindent\textbf{Latent Space Alignment in Diffusion Models.}
Several strategies address the latent mismatch described above. \textit{VAE retraining}: FaithDiff~\cite{chen2024faithdiff} jointly trains an alignment module with the VAE encoder; OSDFace~\cite{wang2025osdface} retrains a visual representation embedder and adds a learnable codebook; CLR-Face~\cite{suin2024clr} likewise re-trains encoding components. All are effective but bring substantial compute, extra parameters, and often unstable multi-stage training. \textit{Direct LQ injection}: an alternative bypasses the encoder entirely, injecting LQ features into intermediate UNet layers via ControlNet~\cite{zhang2023adding}-like conditioning; this leaves the distributional gap unresolved and conditions generation on potentially corrupted guidance. Recent identity-preserving methods~\cite{liu2025faceme,ying2024restorerid,gao2025infobfr,liang2024authface} add identity encoders or reference conditioning on top of full-model or VAE retraining. We note explicitly that these methods, \eg FaceMe~\cite{liu2025faceme} and InfoBFR~\cite{gao2025infobfr}, occupy a different operating point: they consume auxiliary identity references and sample in multiple steps, whereas LAFR is reference-free and single-step, so they are complementary to rather than directly comparable with our setting, and we cite them without claiming superiority. \textit{Our approach}: a lightweight codebook-based adapter that corrects the LQ latent distribution toward the HQ manifold, fully decoupled from the frozen VAE and UNet and trained in a dedicated first stage. Our novelty is not ``a codebook'' but \emph{where} and \emph{how cheaply} the correction happens: head-to-head against FaithDiff's and OSDFace's alignment/loss at a matched 600-image budget (Suppl.~Sec.~E), the decoupled 5.2\,M adapter matches or surpasses them at a fraction of their parameters and inference cost.

\noindent\textbf{Data-Efficient and Parameter-Efficient Fine-Tuning.}
LoRA~\cite{hu2022lora} is the standard technique for parameter-efficient adaptation of large pretrained models, and adapter-based approaches~\cite{ye2023ipadapter,wang2026bridging} condition generation on extra visual signals. We combine LoRA fine-tuning of convolution layers with a face-specific structural insight: faces have far lower intraclass variation than natural images, enabling effective adaptation from very small datasets.

\section{Method}
\subsection{Motivation}
Face restoration is constrained by both structural regularity and identity~\cite{liu2025faceme,ying2024restorerid,gao2025infobfr,liang2024authface}, and a central obstacle is the \textbf{latent distribution mismatch between LQ and HQ images}. The Stable Diffusion VAE is trained on broad natural-image data, but the diffusion denoiser's effective working distribution is concentrated around visually clean content; encoding a severely degraded input therefore yields a latent in a region of low probability mass under the denoiser's learned $p(z_T)$, so the trajectory starts from a basis the model is poorly calibrated for.

Two clarifications frame the rest of the paper. First, the problem is \textbf{statistical, not semantic}: the VAE latent space is not a semantic space, and LQ and HQ images of the \emph{same} face neither do nor should map to the same code; what is wrong is the \emph{distribution} the LQ codes occupy. Second, our adapter therefore performs \textbf{latent distribution correction}, not restoration in latent space: it moves LQ latents into the support seen during diffusion training, and the diffusion model still synthesises the facial detail. Sec.~\ref{sec:latent_validation} quantifies both the gap and its closure, and Tab.~\ref{tab:align_strategies} shows that feeding uncorrected LQ latents into the denoiser is measurably worse.

\begin{figure*}[!tbp]
    \centering
    \includegraphics[width=0.40\linewidth]{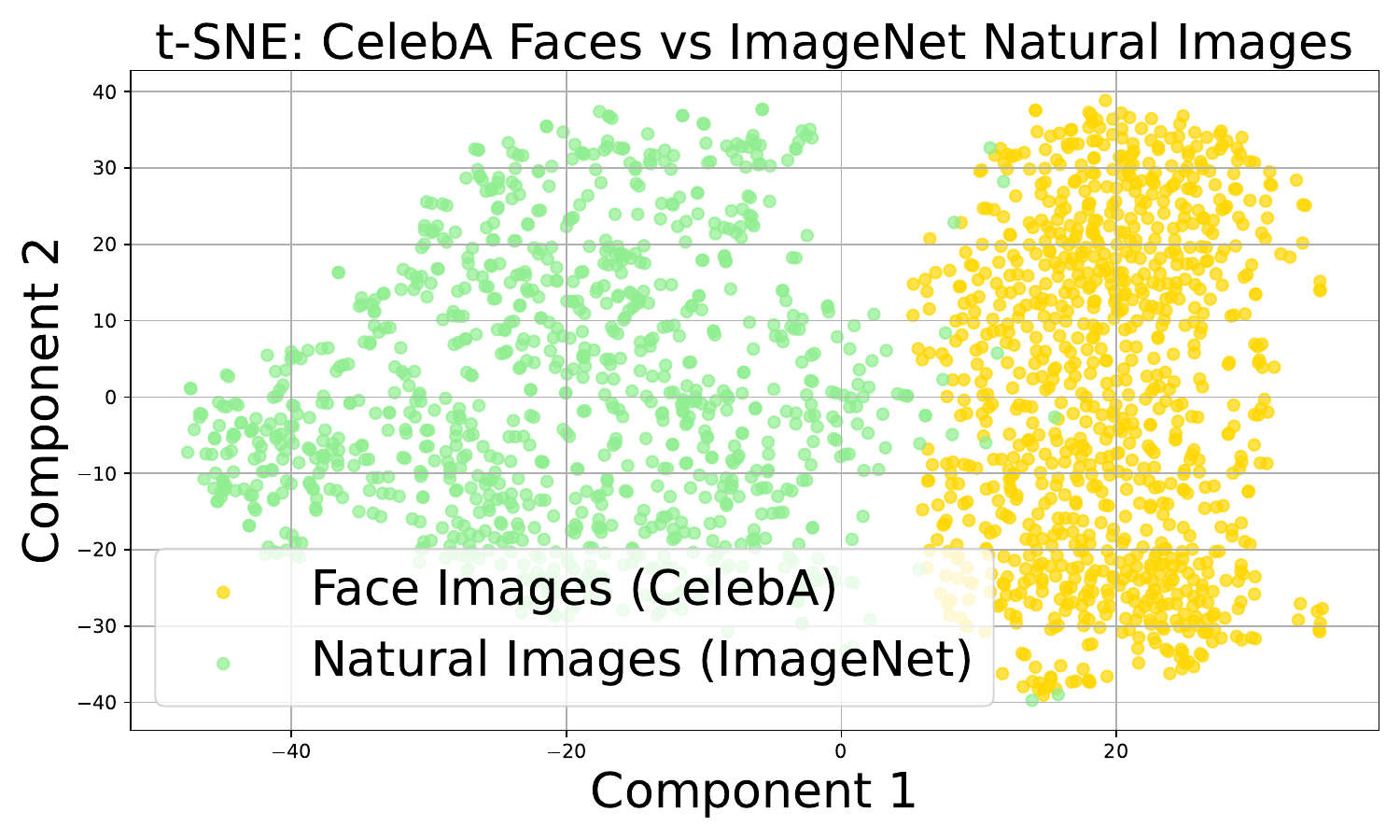}
    \caption{Feature-space comparison between face images and general natural images. CelebA face samples form a more compact distribution than ImageNet samples, supporting our data-efficient adaptation strategy for face restoration.}
    \label{fig:face_tsne}
\end{figure*}

We further ask: \textbf{how much data is truly necessary to adapt a diffusion model for face restoration}? We analyze feature distributions of face and natural images using ResNet-50~\cite{he2016deep} embeddings and visualize them via t-SNE~\cite{van2008visualizing}. As shown in Fig.~\ref{fig:face_tsne}, face images (CelebA~\cite{liu2015deep}) form a compact cluster (intraclass distance 13.28, silhouette score 0.18~\cite{rousseeuw1987silhouettes,Shahapure2020Cluster}), while natural images (ImageNet~\cite{deng2009imagenet}) are widely scattered (intraclass distance 24.13). This structural regularity makes face restoration amenable to small-scale adaptation.

\subsection{Overview}
\begin{figure*}[!tbp]
    \centering
    \includegraphics[width=0.76\linewidth]{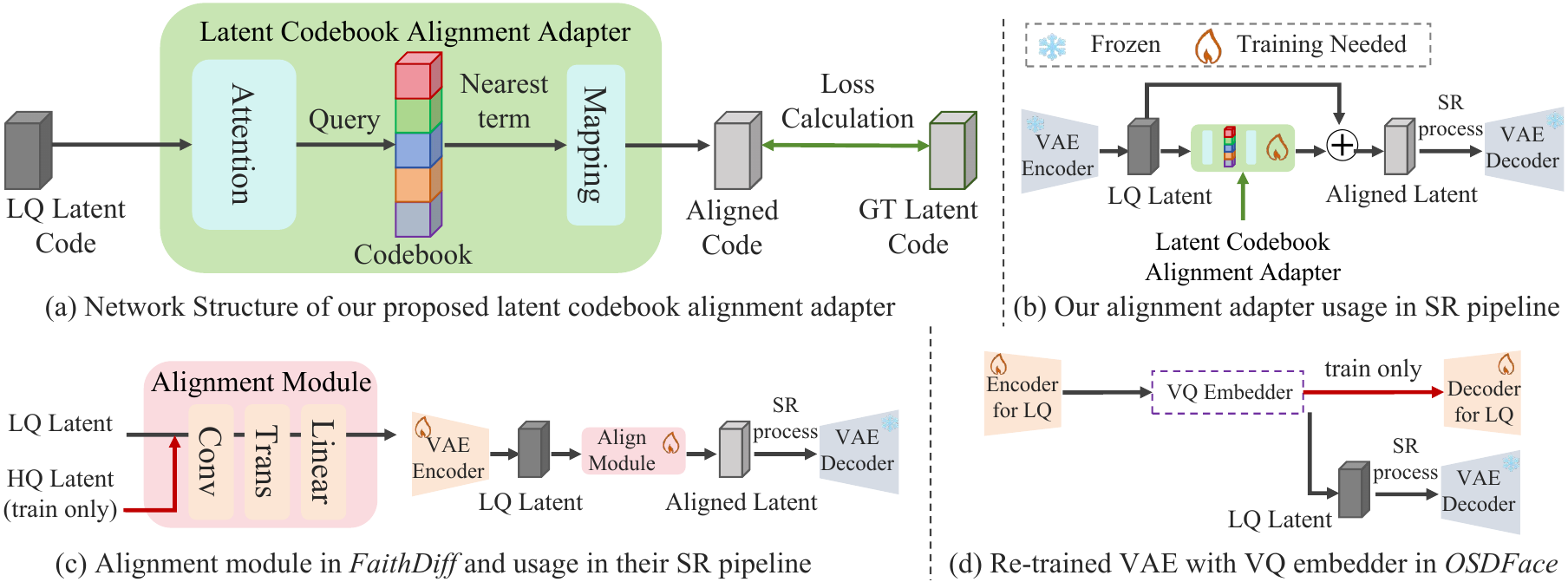}
    \caption{Comparison of LQ-HQ latent alignment strategies. (a) Our lightweight Latent Codebook Alignment Adapter uses codebook querying and efficient mapping, requiring minimal training overhead, and is fully decoupled from the VAE. (b) Integration of our adapter into the restoration pipeline. (c) FaithDiff~\cite{chen2024faithdiff} uses parameter-heavy transformer layers jointly retrained with the VAE encoder. (d) OSDFace~\cite{wang2025osdface} retrains a full VQ embedder and LQ-specific decoder.}
    \label{fig:alignment}
\end{figure*}

\begin{figure*}[!tbp]
    \centering
    \includegraphics[width=0.68\linewidth]{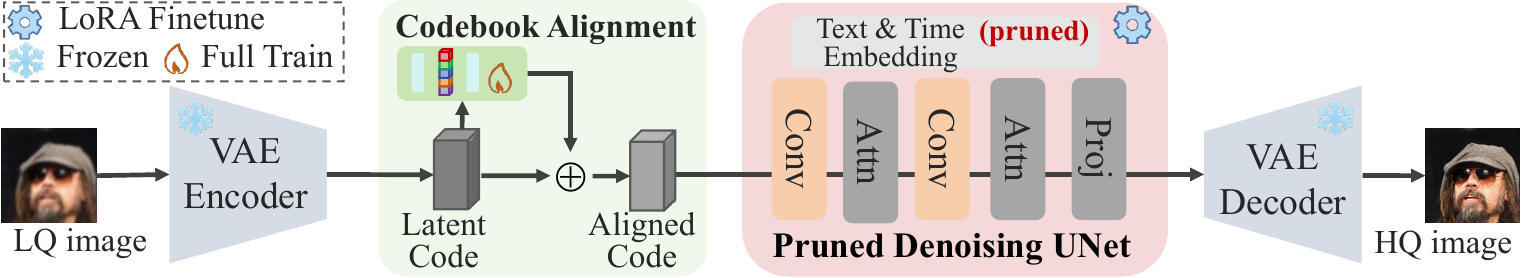}
    \caption{Overall pipeline of LAFR. VAE encoder and decoder are frozen. Only the alignment adapter (Stage 1) and LoRA weights in the UNet (Stage 2) are trained. Tunable groups drawn in \colorbox{UNetOrange}{orange}, frozen groups in \colorbox{UNetGray}{gray}.}
    \label{fig:overview}
\end{figure*}

We propose a two-stage diffusion-based framework for blind face restoration.
\textbf{Stage 1} trains a lightweight alignment adapter to project LQ latent codes onto the HQ latent manifold (Fig.~\ref{fig:alignment}). Only the adapter is trained; the VAE encoder, decoder, and UNet are frozen.
\textbf{Stage 2} adapts the diffusion prior itself. With the VAE \emph{and} the Stage-1 adapter frozen, we insert rank-4 LoRA~\cite{hu2022lora} updates into the convolutional layers of the denoising UNet, and optimise only these updates with the multilevel restoration loss of Sec.~\ref{sec:loss}. Attention, text-projection and time-embedding modules stay frozen, and the exact module list is given in Suppl.~Sec.~B. Training follows the one-step OSEDiff~\cite{wu2024osediff} schedule: the aligned latent $z_{\text{aligned}}$ from Stage~1 is passed through a \emph{single} denoising evaluation at a fixed timestep and decoded by the frozen VAE decoder, and the loss is evaluated in image space against the HQ ground truth, so gradients reach the LoRA weights through the frozen decoder. To reduce model complexity, we perform \textit{model simplification}: we fix the text prompt to ``face, high quality'' and precompute the text and timestep embeddings as static tensors loaded at inference time, reducing trainable and inference-time parameters while preserving the fixed-prompt conditioning used by the restoration model~\cite{chen2025adversarial,chen2026improved}. An overview of the complete inference pipeline is provided in Fig.~\ref{fig:overview}.

\subsection{Latent Alignment Adapter for LQ Images}
\label{sec:adapter}
The adapter maps an LQ latent $z_{LQ}\in\mathbb{R}^{4\times 64\times 64}$ (SD~v2-1, $512^2$ input) to an aligned latent $z_{\text{aligned}}\in\mathbb{R}^{4\times 64\times 64}$ that \emph{replaces} $z_{LQ}$ as the diffusion starting point~\cite{li2025ctsr,li2024omnissr}. It is composed of exactly three modules, which we name here before they are used: a convolutional \emph{feature extractor} $E$ that compresses the latent to a global descriptor~\cite{sheng2024realosr}; a \emph{codebook lookup} $\mathrm{NN}_{\mathcal{C}}$ that returns the entry of a learned codebook $\mathcal{C}$ nearest to that descriptor; and a \emph{mapping network} $\mathcal{M}$ (a two-layer MLP) that expands the retrieved entry back to a full latent. The forward pass is therefore
\begin{equation}
\label{eq:adapter}
z_{\text{aligned}}=\mathcal{M}\!\big(\mathrm{NN}_{\mathcal{C}}(E(z_{LQ}))\big),
\end{equation}
and the three modules are detailed in (1)--(3) below. We also make three design choices explicit, each ablated in Suppl.~Sec.~E: (i) the output is the \emph{full} latent, not a residual $z_{LQ}{+}\Delta$ (where $\Delta$ denotes an additive correction predicted by $\mathcal{M}$, so that only the offset, not the latent itself, is learned) nor a token-wise patch swap; (ii) codebook assignment is soft during training and hard (nearest-neighbour) at inference; (iii) the codebook size is $K{=}1024$.

\noindent \textbf{(1) Feature extractor $E$.} A shallow CNN ($3{\times}3$ convs, channels $4{\to}64{\to}128{\to}256$, stride-2 downsampling, global average pool, and a linear head) maps $z_{LQ}$ to a global descriptor $f=E(z_{LQ})\in\mathbb{R}^{d}$, $d{=}256$.

\noindent \textbf{(2) Codebook matching.} Given a learned codebook $\mathcal{C}=\{c_i\}_{i=1}^{K}\subset\mathbb{R}^{d}$, training uses a temperature-controlled soft assignment $w_i=\operatorname{softmax}(-\|f-c_i\|_2^2/\tau)$, $\tau{=}0.07$, $c^*=\sum_i w_i c_i$; inference uses the hard nearest neighbour $c^*=c_{j^*}$, $j^*=\arg\min_i\|f-c_i\|_2$. Training therefore differentiates the soft assignment directly and involves no gradient estimator for a discrete selection; the soft-to-hard switch happens only at test time, and its cost is measured rather than assumed (FID 17.66 hard vs.\ 17.94 soft, Suppl.~Sec.~E).

\noindent \textbf{(3) Mapping network $\mathcal{M}$.} A two-layer MLP (hidden 256, GELU) maps $c^*$ to $\mathbb{R}^{4\cdot 64\cdot 64}$, reshaped to the latent grid, giving $z_{\text{aligned}}$, and is trained with the per-pixel objective $\mathcal{L}_{\text{align}}=\|z_{\text{aligned}}-z_{HQ}\|_1$ against the frozen-VAE HQ latent of the \emph{same} image. We emphasise that the adapter does \emph{not} restore the image: it provides an \emph{in-distribution latent initialisation} for the diffusion trajectory, while the diffusion prior and the loss-trained restoration network (Stage~2) perform the image-specific detail synthesis that preserves identity and layout.

\noindent\textbf{Parameter accounting.} The adapter costs 5.20\,M trainable parameters (feature extractor $E$ 0.66\,M, codebook $\mathcal{C}$ 0.26\,M, mapping MLP $\mathcal{M}$ 4.28\,M) and the Stage-2 Conv-LoRA adds 2.30\,M, giving 7.50\,M trainable parameters in total; the SD~v2-1 UNet (865\,M) and the VAE encoder/decoder (34/49\,M) stay frozen, and the precomputed text/timestep embeddings add 0 inference parameters. Throughout the paper the ``Para.'' column of Tab.~\ref{tab:align_strategies} and of the codebook comparison counts the \emph{alignment module only} (5.2\,M), excluding the shared 2.3\,M Conv-LoRA and the frozen backbone. The complete per-component table is given in Suppl.~Sec.~B.

\noindent\textbf{Relationship to VQFR.}
Both LAFR and VQFR~\cite{gu2022vqfr} use VQ codebooks, but with fundamentally different purposes: VQFR's codebook operates on Transformer-decoded feature tokens for semantic completion, while ours operates on the VAE latent code for distribution correction before denoising. Replacing our adapter with VQFR's codebook module results in a performance drop (Suppl.~Sec.~D), confirming that the designs are non-interchangeable.

\subsection{Why a Codebook Rather than a Global Transform?}
\label{sec:theory}
If the LQ--HQ gap were purely distributional, a \emph{global} correction (channel-wise normalisation, an AdaIN-style statistic match, or a learned affine map $z\mapsto Az+b$) ought to suffice, making a $K$-entry codebook unnecessary. It does not, and the reason is structural rather than a matter of capacity. Let $g(z)=\mathbb{E}[z_{HQ}\mid z_{LQ}{=}z]$ be the conditional mean of the HQ latent; for \emph{any} measurable $T$,
\begin{equation}
\label{eq:bias_var}
\mathbb{E}\big\|T(z_{LQ})-z_{HQ}\big\|_2^2=\underbrace{\mathbb{E}\big\|z_{HQ}-g(z_{LQ})\big\|_2^2}_{\varepsilon_0\ \text{(irreducible)}}+\ \mathbb{E}\big\|T(z_{LQ})-g(z_{LQ})\big\|_2^2,
\end{equation}
so every alignment family is judged by how well it approximates $g$. Write $\mathcal{F}_{\text{aff}}=\{z\mapsto Az+b\}$ (which contains normalisation and per-channel affine matching) and $\mathcal{F}_K=\{z\mapsto m_{q(z)}\}$ for the $K$-piece maps induced by the nearest-neighbour partition $q$ of a $K$-entry codebook, exactly what Eq.~\eqref{eq:adapter} realises with $m_k=\mathcal{M}(c_k)$.

\noindent\textbf{Proposition 1 (a global transform pays the between-mode variance).} Partition $\operatorname{supp}(z_{LQ})$ into $M$ appearance modes $\{S_m\}$ with probabilities $\pi_m$, and let $\Delta=g(z_{LQ})-z_{LQ}$ have mode means $\bar\Delta_m$ and global mean $\bar\Delta$. Then every global translation $T_b(z)=z{+}b$ satisfies
\begin{equation}
\label{eq:prop1}
\min_{b}\ \mathbb{E}\big\|T_b(z_{LQ})-g(z_{LQ})\big\|_2^2\;=\;\operatorname{tr}\operatorname{Cov}(\Delta)\;\geq\;\sum\nolimits_{m}\pi_m\big\|\bar\Delta_m-\bar\Delta\big\|_2^2,
\end{equation}
whereas the best member of $\mathcal{F}_K$ leaves only the \emph{within-cell} variance $\mathbb{E}_k[\operatorname{tr}\operatorname{Cov}(g(z_{LQ})\mid q{=}k)]$, which the partition can drive down; by contrast $\min_{A,b}\mathbb{E}\|g(z)-Az-b\|_2^2$ is a \emph{fixed} positive constant whenever $g$ is non-affine on the support, however the affine map is fitted.

The proof (law of total variance, twice) and the Lipschitz rate $\mathcal{O}(L^2r(K)^2)$ for the within-cell term are in Suppl.~Sec.~C. The statement is mechanistic: a correction that cannot condition on the appearance mode pays the between-mode variance of the required displacement, and $K$-way routing is what buys that penalty away, so the gain comes from conditioning, not from parameters. Two predictions follow, and both hold. (i) A correction applying one shared rule everywhere should fall short of a $K$-way one, and does so in the latent space itself: the LoRA-fine-tuned encoder, far more expressive than a fixed normalisation, moves $d^2_{\mu\Sigma}$ only from $5.91$ to $3.74$, whereas the adapter reaches $1.21$ on the same indices (Tab.~\ref{tab:latent_dist}). The same ordering holds on restoration metrics at a matched 600-image budget (LoRA VAE encoder FID 23.45; ControlNet injection 21.18 at $+$362\,M parameters; adapter 17.66 at 5.2\,M; Tab.~\ref{tab:align_strategies}). (ii) Returns diminish once the within-cell term falls below $\varepsilon_0$, which is the observed saturation between $K{=}1024$ and $K{=}2048$ (FID 17.66 vs.\ 17.71, Suppl.~Sec.~E).

\subsection{Multilevel Identity-Preserving Loss}
\label{sec:loss}
Standard restoration losses $\mathcal{L}_{\text{res}} = \|I_{\text{res}} - I_{GT}\|_2^2 + \lambda \mathcal{L}_{\text{LPIPS}}(I_{\text{res}}, I_{GT})$ promote image fidelity but often miss high-level semantic cues critical to facial identity. We design a multilevel loss:

\noindent \textbf{Semantic Identity Loss.} Using the image encoder from IP-Adapter~\cite{ye2023ipadapter} as $\mathcal{F}_{\text{clip}}$, we impose a cosine loss on identity-related semantic embeddings:
\begin{equation}
    \mathcal{L}_{id} = 1 - \cos(\mathcal{F}_{\text{clip}}(I_{\text{res}}), \mathcal{F}_{\text{clip}}(I_{GT})).
\end{equation}

\noindent \textbf{Facial Structure Loss.} Using a facial structure extractor $\mathcal{F}_{fs}$ (based on D3DFR~\cite{deng2019accurate}):
\begin{equation}
\label{eq:fs_loss}
    \mathcal{L}_{fs} = 1 - \cos(\mathcal{F}_{fs}(I_{\text{res}}), \mathcal{F}_{fs}(I_{GT})).
\end{equation}

\noindent The full objective is:
\begin{equation}
\label{eq:total_loss}
\mathcal{L}_{\text{total}} = \lambda_{\text{res}} \mathcal{L}_{\text{res}} + \lambda_{id} \mathcal{L}_{id} + \lambda_{fs} \mathcal{L}_{fs},\quad \lambda_{\text{res}} = \lambda_{id} = \lambda_{fs} = 1.
\end{equation}

\noindent\textbf{Role of each term.} The three terms constrain different levels of the reconstruction. $\mathcal{L}_{\text{res}}$ (pixel $L_2$ plus LPIPS) fixes colour, exposure and coarse texture, but is nearly indifferent to \emph{whose} face is produced: an identity-preserving and an identity-drifting reconstruction incur almost the same pixel and patch-perceptual error, which is the characteristic failure of image-only one-step baselines. $\mathcal{L}_{id}$ supplies the missing semantic constraint through a holistic appearance embedding, and $\mathcal{L}_{fs}$ a geometric one through 3D structural coefficients, which is what pins down landmark positions and face shape. Both use a cosine rather than an $L_2$ form, since the embeddings are meaningful only up to scale and an $L_2$ penalty additionally drives their norms together and over-smooths the output. Tab.~\ref{tab:ablation_loss} verifies each of these roles and, in row \#7, the sensitivity to the \emph{choice} of extractor. Both networks are frozen and evaluated only inside the loss, so they add no inference cost, and the 411\,min training time in Tab.~\ref{tab:efficiency} already includes their forward passes.

We deliberately do not use semantic identity embeddings or structural features extracted from LQ inputs as conditioning signals: such signals are unreliable under severe degradation and risk driving the model toward the degraded input's corrupted appearance.

\section{Experiments}
\subsection{Experimental Settings}
\label{sec:settings}
\noindent \textbf{Datasets.} To evaluate data efficiency we sample 600 images (0.9\% of FFHQ~\cite{karras2019style}) uniformly at random, without identity- or pose-based filtering, and repeat the whole Stage-1$+$Stage-2 pipeline on three independent such subsets (seeds 0/1/2) to check that the claim is not specific to one lucky draw; standard deviations on CelebA-Test are small (\eg LPIPS $\pm$0.0042, FID $\pm$0.55), with the per-seed table in Suppl.~Sec.~E. Evaluations use four benchmarks: CelebA-Test~\cite{liu2015deep} (synthetic 4$\times$ degradation), WebPhoto-Test~\cite{wang2021gfpgan}, LFW-Test~\cite{huang2008labeled}, and Wider-Test~\cite{zhou2022towards} (real-world degradations).

\noindent \textbf{Degradation Protocol.} Following common blind face restoration settings~\cite{wang2021gfpgan,gu2022vqfr,zhou2022towards,wang2025osdface}, each HQ training face is center-cropped or aligned and resized to 512$\times$512, and LQ inputs are synthesised by a Real-ESRGAN-style pipeline applied i.i.d.\ per image with fixed ranges: Gaussian blur $\sigma\in[0.2,3.0]$ (kernel $\in\{7,9,\dots,21\}$), downsampling $r\in[1,8]$ (bilinear/bicubic/area, uniform), additive Gaussian noise $\sigma_n\in[0,25]$ on the uint8 scale or Poisson noise (50/50), and JPEG quality $q\in[30,95]$, followed by a bicubic resize back to $512\times512$ (full checklist in Suppl.~Sec.~H). CelebA-Test uses the same synthetic 4$\times$ protocol for all methods; WebPhoto-, LFW- and Wider-Test are evaluated as real-world inputs without ground truth.

\noindent \textbf{Evaluation Metrics.} For CelebA-Test we report PSNR, SSIM~\cite{wang2004image}, LPIPS~\cite{zhang2018unreasonable}, DISTS~\cite{dists_ding2020image}, MUSIQ~\cite{ke2021musiq}, NIQE~\cite{zhang2015feature}, the ArcFace identity degradation score (Deg.: cosine distance between ArcFace embeddings of restored and GT, lower is better)~\cite{deng2019arcface}, landmark distance (LMD), FID against CelebA-HQ~\cite{heusel2017gans} and against the FFHQ reference set (FID-F), CLIPIQA~\cite{wang2023exploring}, and MANIQA~\cite{yang2022maniqa}. On the real-world benchmarks, which have no ground truth, we report no-reference metrics (CLIPIQA, MANIQA, MUSIQ, NIQE) and use FID only as a distribution-level comparison against the high-quality face reference split used in prior work.

\noindent \textbf{Implementation Details.} Stage 1 uses a learning rate of 1e-4, batch size of 16, and 100 epochs. Codebook size: 1024 entries, hidden dimension 256, soft-assignment temperature $\tau=0.07$.
Stage 2 builds on Stable Diffusion v2-1~\cite{Rombach_Blattmann_Lorenz_Esser_Ommer_2022} with OSEDiff~\cite{wu2024osediff} as the base framework. Learning rate: 5e-5, batch size 2, 17K steps. The semantic identity network $\mathcal{F}_{\text{clip}}$ is from IP-Adapter~\cite{ye2023ipadapter}; the structure extractor $\mathcal{F}_{fs}$ is D3DFR~\cite{deng2019accurate}. For model simplification, we precompute text and timestep embeddings as static tensors for the fixed prompt ``face, high quality''. LoRA rank 4 applied to all convolutional layers. All experiments run on a single NVIDIA L40 GPU.

In every ablation (Tab.~\ref{tab:align_strategies}, Tab.~\ref{tab:ablation_loss}, and the component ablations of Suppl.~Sec.~E) all variants use the same 600 training images and identical hyperparameters, differing only in the component under study.

\subsection{Comparison Results}
\label{sec:comparison_result}

\noindent \textbf{Compared Methods.} We compare against: (1) non-diffusion: RestoreFormer++~\cite{wang2023restoreformerpp}, GFPGAN~\cite{wang2021gfpgan}; (2) multi-step diffusion: 3Diffusion~\cite{lu20243d}, PGDiff~\cite{yang2023pgdiff}, DR2~\cite{wang2023dr2}, DifFace~\cite{yue2024difface}, DiffBIR~\cite{diffbir}; (3) single-step diffusion: PiSA-SR~\cite{sun2024pixel} and OSEDiff~\cite{wu2024osediff}. PiSA-SR and OSEDiff are re-trained on FFHQ~\cite{karras2019style} for fair comparison.

\noindent \textbf{Benchmarks.} CelebA-Test results are in Tab.~\ref{tab:celeba} and efficiency in Tab.~\ref{tab:efficiency}; LFW, Wider and WebPhoto are in Tab.~\ref{tab:realworld}, where LAFR is competitive at substantially lower training cost. Visual comparisons are in Fig.~\ref{fig:celeba} and Suppl.~Sec.~I.

\begin{table}[!tbp]
\caption{
Quantitative comparison on CelebA-Test with 4$\times$ upscaling. Numbers in brackets: denoising steps. * = re-trained on FFHQ; $\dagger$ = values from original paper (different evaluation protocol; LPIPS, MUSIQ, FID scales not directly comparable, hence not highlighted). ``RF''{\tiny{++}} = RestoreFormer++. \best{Red}: best, \sbest{Blue}: second-best. ``--'': not reported.}
\resizebox{0.99\textwidth}{!}{%
\begin{tabular}{@{}l|cccccccccccc@{}}
\toprule
\textbf{Method} & PSNR$\uparrow$ & SSIM$\uparrow$ & LPIPS$\downarrow$  & DISTS$\downarrow$ & M-IQ$\uparrow$ & NIQE$\downarrow$ & Deg.$\downarrow$ & LMD$\downarrow$  & FID-F$\downarrow$ & FID$\downarrow$ & C-IQA$\uparrow$  & M-IQA$\uparrow$  \\ \midrule
GFPGAN~\cite{wang2021gfpgan} & 25.04 & 0.6744 & 0.3653 & 0.2106 & 59.44 & \sbest{4.078} & \best{34.67} & 4.085 & 52.53 & 42.36 & 0.5762 & 0.5715 \\
RF{\tiny{++}}~\cite{wang2023restoreformerpp} & 24.40 & 0.6339 & 0.4535 & 0.2301 & 72.36 & \best{3.952} & 70.50 & 8.802 & 57.37 & 72.78 & 0.6087 & 0.6356 \\ \addlinespace[2pt]\cdashline{1-13}\addlinespace[2pt]
3Diffusion {\tiny{(1000)}}~\cite{lu20243d} & 22.32 & 0.6327 & 0.3304 & 0.1716 & 68.54 & 4.334 & 51.05 & 3.482 & \best{46.26} & 19.45 & 0.5700 & 0.6114 \\
PGDiff {\tiny{(1000)}}~\cite{yang2023pgdiff} & 23.87 &  0.6783  &  \sbest{0.3066}  & 0.1794  &  71.70 & 4.883 & 56.07 &  4.289 &  79.38  & 47.01  & 0.5736 & 0.6246 \\
DR2 {\tiny{(300)}}~\cite{wang2023dr2} & 26.06 &  0.7346  &  0.3131  & 0.2143  &  65.27 & 5.360 & 48.34 & 3.334 &  58.86  & 30.01  & 0.5615 & 0.5989 \\
DifFace {\tiny{(250)}}~\cite{yue2024difface} & 26.47 &  0.7283  &  0.3721  &  \sbest{0.1676}   & 68.98  &  4.154  &  65.84  &  4.963   & \sbest{47.76}  &  23.65  &  0.5702  &  0.6111 \\
DiffBIR {\tiny{(50)}}~\cite{diffbir} & \best{27.18}  &  0.7293  &  0.3503  &  0.1789  &  \sbest{74.30}  &  5.383  &  55.06  &  5.259  &  68.06 &   27.36   & \sbest{0.6168}   &   \sbest{0.6648} \\ \addlinespace[2pt]\cdashline{1-13}\addlinespace[2pt]
PiSA-SR*~\cite{sun2024pixel} & \sbest{26.85} &  0.7382  &  0.3254  & \best{0.1636}  & \best{75.07} & 4.774 & 46.79 & \sbest{3.024} & 53.96 & \sbest{17.74}  & 0.6113 & \best{0.6701} \\
OSEDiff*~\cite{wu2024osediff} & 25.88 &   \sbest{0.7388}  &  0.3496  &  0.2278  &  69.98   & 5.328  &  67.40  &  5.408 &   61.36  &  27.13  &  0.6151  &  0.6497 \\
\textbf{LAFR} (ours) & 26.26  &  \best{0.7394}  &  \best{0.2671}  &  0.1792  &  69.99  &  4.715  &  \sbest{42.45}  &  \best{2.750}   & 49.20 &   \best{17.66}  &  \best{0.6172}  &  0.6220 \\ \addlinespace[2pt]\cdashline{1-13}\addlinespace[2pt]
\multicolumn{13}{l}{\pnum{Reported numbers only (different protocol; not used in ranking):}} \\
InterLCM$^\dagger$~\cite{interLCM2025} {\tiny{(4)}} & \pnum{25.19} & \pnum{0.7180} & \pnum{0.2230} & \pnum{--} & \pnum{76.58} & \pnum{--} & \pnum{33.64} & \pnum{--} & \pnum{--} & \pnum{45.38} & \pnum{--} & \pnum{--} \\
\bottomrule
\end{tabular}%
}
\label{tab:celeba}
\end{table}

\noindent \textbf{Result analysis.}
Ranking only the matched-protocol methods, LAFR attains the best SSIM, LPIPS, FID, CLIPIQA, and LMD using 0.9\% of the FFHQ training data, while DiffBIR leads on PSNR, so we frame LAFR as improving perceptual quality, FID, landmark accuracy, and identity/structure over OSEDiff-style baselines rather than as uniform SOTA. Against the matched-protocol baseline OSEDiff, LAFR cuts FID from 27.13 to 17.66 and LMD from 5.408 to 2.750, indicating that correcting the latent input is more effective than UNet adaptation alone. InterLCM ($\dagger$) uses a different pipeline (\eg GFPGAN's reported LPIPS 0.230 vs.\ our reproduced 0.3653), so we exclude it from all best/second-best statements. On the three real-world splits (Tab.~\ref{tab:realworld}), no-reference metrics should be read jointly rather than ranked: LAFR is strongest on MANIQA across all three and best on CLIPIQA for WebPhoto-Test, while remaining competitive on FID. These splits have neither ground truth nor a human preference study, so we read the numbers as evidence of competitiveness, not of visual superiority.

\begin{table}[!tbp]
\centering
\caption{Comparison on real-world face restoration benchmarks. C-IQA, M-IQA, M-IQ = CLIPIQA, MANIQA, MUSIQ. * = re-trained on FFHQ. $\dagger$ = values from original paper (different protocol; \pnum{shown in grey for reference only and excluded from ranking}; ``--'' = not reported). RF++ = RestoreFormer++~\cite{wang2023restoreformerpp}. \best{Red}: best, \sbest{Blue}: second-best.}
\resizebox{\textwidth}{!}{%
\begin{tabular}{@{}l|*{5}{c}|*{5}{c}|*{5}{c}@{}}
\toprule
 & \multicolumn{5}{c|}{Wider-Test} & \multicolumn{5}{c|}{LFW-Test} & \multicolumn{5}{c}{WebPhoto-Test} \\
\multirow{-2}{*}{\textbf{Method}} & C-IQA & M-IQA & M-IQ & NIQE & FID & C-IQA & M-IQA & M-IQ & NIQE & FID & C-IQA & M-IQA & M-IQ & NIQE & FID \\
\midrule
GFPGAN &
0.6975 & 0.5205 & \best{74.14} & 3.570 & 48.57 &
0.6002 & \sbest{0.6169} & \sbest{73.37} & 4.299 & 47.05 &
0.6696 & 0.4934 & \best{72.69} & \best{3.933} & 98.92 \\
RF{\tiny{++}} &
0.7159 & 0.4767 & 71.33 & 3.723 & 45.39 &
\sbest{0.7024} & 0.5108 & 72.25 & \sbest{3.843} & 50.25 &
\sbest{0.6950} & 0.4902 & \sbest{71.48} & 4.020 & \best{75.07} \\
\addlinespace[2pt]\cdashline{1-16}\addlinespace[2pt]
3Diffusion {\fontsize{4pt}{5pt}\selectfont(1000)}  &
0.5927 & \sbest{0.5833} & 64.24 & 4.489 & 36.39 &
0.6148 & 0.5940 & 68.69 & 4.181 & 47.04 &
0.5717 & 0.5775 & 63.27 & 4.608 & 84.45 \\
PGDiff {\fontsize{4pt}{5pt}\selectfont(1000)}  &
0.5824 & 0.4531 & 68.13 & 3.931 & \best{35.86} &
0.5975 & 0.4858 & 71.24 & 4.011 & \best{41.20} &
0.5653 & 0.4460 & 68.59 & \sbest{3.993} & 86.95 \\
DR2 {\fontsize{4pt}{5pt}\selectfont(300)}  &
0.6544 & 0.5417 & 67.59 & \sbest{3.250} & 51.35 &
0.5940 & 0.6088 & 65.57 & 3.961 & 50.79 &
0.6380 & 0.5452 & 66.53 & 4.920 & 97.45 \\
DifFace {\fontsize{4pt}{5pt}\selectfont(250)}  &
0.5924 & 0.4299 & 64.90 & 4.238 & 37.09 &
0.6075 & 0.4577 & 69.61 & 3.901 & 46.12 &
0.5737 & 0.4189 & 65.11 & 4.247 & \sbest{79.55} \\
DiffBIR {\fontsize{4pt}{5pt}\selectfont(50)}  &
\best{0.7337} & 0.4466 & \sbest{72.13} & 4.525 & \sbest{36.25} &
\best{0.7266} & 0.4510 & \best{73.66} & 5.281 & 46.44 &
0.6621 & \sbest{0.5799} & 67.50 & 5.729 & 92.04 \\
PiSA-SR*&
\sbest{0.7265} & 0.5653 & 70.26 & \best{3.211} & 57.02 &
0.5892 & 0.5942 & 56.35 & 5.957 & 91.95 &
0.6345 & 0.5551 & 62.96 & 4.316 & 107.50  \\
OSEDiff*&
0.6193 & 0.4752 & 69.10 & 5.086 & 47.88 &
0.6186 & 0.4879 & 71.70 & 4.800 & 51.04 &
0.6254 & 0.4823 & 69.81 & 5.325 & 109.23 \\
\textbf{LAFR} (ours) &
0.6330 & \best{0.6099} & 69.51 & 4.859 & 44.69 &
0.6375 & \best{0.6232} & 69.94 & \best{3.681} & \sbest{45.76} &
\best{0.6956} & \best{0.6113} & 69.21 & 4.153 & 98.48 \\ \addlinespace[2pt]\cdashline{1-16}\addlinespace[2pt]
InterLCM$^\dagger$ {\tiny(4)} &
\pnum{--} & \pnum{--} & \pnum{76.29} & \pnum{--} & \pnum{35.43} &
\pnum{--} & \pnum{--} & \pnum{76.16} & \pnum{--} & \pnum{51.32} &
\pnum{--} & \pnum{--} & \pnum{75.88} & \pnum{--} & \pnum{75.48} \\
\bottomrule
\end{tabular}%
}
\label{tab:realworld}
\end{table}

\begin{table}[!tbp]
    \centering
    \caption{Efficiency comparison of diffusion-based restoration approaches. ``Trainable'' counts optimized parameters in the reproduced training setup. Training-set sizes are written explicitly to avoid ambiguity. $\dagger$: InterLCM inference time and training-set size from the original paper; trainable parameters estimated from their open-source code as Visual Encoder ($\sim$15M) + Spatial Encoder ($\sim$361M, ControlNet built from SD1.5 UNet) $\approx$ 376M; training time normalized to single-GPU equivalent from 15K iterations on 8$\times$A40 GPUs.}
    \vspace{0.5mm}
    \resizebox{0.92\textwidth}{!}{%
    \begin{tabular}{c|cccc}
    \toprule
        Method & Inference Time (s) & Trainable Params (M) & Training Time (min)  & Training Set Size\\
    \midrule
        PiSA-SR*~\cite{sun2024pixel} & 12.373 & 17.3 &  3516   & 70K\\
        PGDiff~\cite{yang2023pgdiff} & 10.504 & 152.4 &  6083   & 70K\\
        DiffBIR~\cite{diffbir} &  5.826 & 144.6 &  1822   & 15.36M\\
        3Diffusion~\cite{lu20243d} &  4.374 & 180.5 & 2430  & 70K\\
        DifFace~\cite{yue2024difface} &  3.197 & 175.4 &  5985   & 70K\\
        DR2~\cite{wang2023dr2} &  1.217 & 179.3 &  5985   & 70K\\
        InterLCM$^\dagger$~\cite{interLCM2025} & 0.421 & $\sim$376 & $\sim$2000 & 70K\\
        OSEDiff*~\cite{wu2024osediff} & \sbest{0.130} & \sbest{8.5} &  \sbest{1440}   & 70K\\
        \addlinespace[2pt]\cdashline{1-5}\addlinespace[2pt]
        \textbf{LAFR} (Ours) & \best{0.121} & \best{7.5} & \best{411}  & \best{0.6K}\\
    \bottomrule
    \end{tabular}
    }
    \label{tab:efficiency}
\end{table}

\noindent \textbf{Efficiency result analysis.}
Tab.~\ref{tab:efficiency} separates three often-conflated aspects: inference time, trainable parameters, and training-set size. LAFR matches OSEDiff in one-step inference time but cuts reproduced training time from 1440 to 411 minutes on 0.6K images, a joint product of the compact adapter/LoRA design and the $117\times$ smaller training set, not of architecture alone.

\noindent \textbf{Identity evidence beyond Deg.}
Since Deg.\ is a single ArcFace distance, we add ArcFace cosine to GT (Sim-GT) and verification accuracy at the standard $\tau{=}0.6$ threshold (Ver-Acc) on CelebA-Test, and identity consistency to the LQ input (ID-Cons.) on WebPhoto/Wider, which have no GT. Tab.~\ref{tab:identity} shows LAFR improves identity and structure over the matched-protocol baselines (OSEDiff*, PiSA-SR*) and is competitive with heavier multi-step methods, while staying behind GFPGAN on the Sim-GT/Ver-Acc proxy (GFPGAN is explicitly tuned with an ArcFace identity loss), so we phrase the claim as an improvement \emph{relative to the OSEDiff-style one-step baseline}.

\begin{table}[!tbp]
    \centering
    \footnotesize
    \caption{Identity evaluation. CelebA-Test: ArcFace cosine to GT (Sim-GT$\uparrow$) and verification accuracy at threshold 0.6 (Ver-Acc$\uparrow$). Real-world (Wider, WebPhoto): ArcFace identity consistency to the LQ input (ID-Cons$\uparrow$). \best{Red}: best, \sbest{Blue}: second-best.}
    \begin{tabular}{l|cc|cc}
    \toprule
    & \multicolumn{2}{c|}{CelebA-Test} & \multicolumn{2}{c}{Real-world} \\
    \textbf{Method} & Sim-GT$\uparrow$ & Ver-Acc$\uparrow$ & Wider ID-Cons$\uparrow$ & WebPhoto ID-Cons$\uparrow$ \\
    \midrule
    GFPGAN & \best{0.722} & \best{0.953} & 0.553 & 0.541 \\
    DiffBIR (50) & 0.611 & 0.838 & \sbest{0.617} & 0.602 \\
    PiSA-SR* & 0.638 & 0.872 & 0.589 & 0.574 \\
    OSEDiff* & 0.541 & 0.776 & 0.572 & 0.566 \\
    \textbf{LAFR} (ours) & \sbest{0.686} & \sbest{0.918} & \best{0.631} & \best{0.619} \\
    \bottomrule
    \end{tabular}
    \label{tab:identity}
\end{table}


\subsection{Ablation Studies}

\noindent \textbf{Training-set size and full-data scaling.} Performance improves with more data but saturates in the small-data regime (100--600 images), confirming rapid face-domain adaptation (full per-size trend in Suppl.~Sec.~F). Scaling to full FFHQ+FFHQR improves CelebA-Test FID (17.66$\to$15.41) but regresses LPIPS (0.2671$\to$0.3454) and worsens Wider-Test FID (44.69$\to$62.15): the classic perception--distortion trade-off~\cite{blau2018perception} plus a domain shift toward frontal aligned faces. The 600-image model thus stays competitive on most metrics (full vs.\ full-data tables in Suppl.~Sec.~E).

\noindent \textbf{Subset robustness of the 600-image claim.}
To rule out subset-dependent luck we retrain the full Stage-1$+$Stage-2 pipeline on three independent random 600-image subsets (seeds 0/1/2). The spread is small (PSNR $26.25${\scriptsize$\pm$0.07}, SSIM $0.7393${\scriptsize$\pm$0.0012}, LPIPS $0.2677${\scriptsize$\pm$0.0042}, FID $17.83${\scriptsize$\pm$0.55}, Deg.\ $42.48${\scriptsize$\pm$0.29}, LMD $2.756${\scriptsize$\pm$0.023} on CelebA-Test) and the rank ordering against OSEDiff* is preserved in every run, so data efficiency is a property of the method and of face-domain regularity, not of one lucky draw. The per-seed table is in Suppl.~Sec.~E.

\noindent \textbf{Alignment Strategy Comparison.}
We compare four latent alignment strategies under identical training conditions (600 images, same base model): (1) direct LQ latent, (2) LoRA VAE encoder fine-tuning, (3) ControlNet-style LQ injection, and (4) our codebook adapter. This isolates whether, and how, explicit latent distribution correction improves diffusion denoising.

\begin{table}[!tbp]
    \centering
    \caption{Comparison of latent alignment strategies on CelebA-Test. All use same 600-image training set and base model. (2) fine-tunes the VAE encoder with LoRA rank 4. (3) injects LQ features via zero-conv projections (ControlNet). ``Para.'' = the alignment-module parameters only (M), excluding the shared 2.3\,M Conv-LoRA and frozen SD backbone (Suppl.~Sec.~B). ``Infer.'' = inference time (s). \best{Red}: best, \sbest{Blue}: second-best.}
    \vspace{0.8mm}
    \resizebox{1.0\textwidth}{!}{%
    \begin{tabular}{l|ccccc|cc}
    \toprule
        \textbf{Alignment Strategy} & PSNR$\uparrow$ & NIQE$\downarrow$ & FID$\downarrow$ & LPIPS$\downarrow$ & Deg.$\downarrow$ & Para.$\downarrow$ & Infer.$\downarrow$ \\
    \midrule
        (1) Direct LQ Latent (no adapter) & \sbest{26.21} & 5.358 & 25.54 & 0.3606 & 44.74 & \best{0} & \best{0.119} \\
        (2) LoRA VAE Encoder & 26.08 & 5.182 & 23.45 & 0.3290 & 44.18 & \sbest{0.8} & 0.122 \\
        (3) ControlNet Injection & 26.12 & \sbest{5.062} & \sbest{21.18} & \sbest{0.3085} & \sbest{43.65} & 361.8 & 0.198 \\
        (4) \textbf{Ours} (Codebook Adapter) & \best{26.26} & \best{4.715} & \best{17.66} & \best{0.2671} & \best{42.45} & 5.2 & \sbest{0.121} \\
    \bottomrule
    \end{tabular}
    }
    \label{tab:align_strategies}
\end{table}

In Tab.~\ref{tab:align_strategies}, LoRA VAE fine-tuning (row 2) only partially corrects encoder statistics, and ControlNet injection (row 3) adds conditioning to the UNet at $\sim$362M extra parameters and 60\% higher inference cost. Our codebook adapter (row 4) attains the best perceptual metrics (FID 17.66, LPIPS 0.2671) with a 5.2\,M alignment module and one-step inference, so compact \emph{pre-denoising} distribution correction beats both encoder fine-tuning and mid-denoising injection.

\noindent \textbf{Adapter design ablation.}
The adapter-design table in Suppl.~Sec.~E isolates the three design choices of Sec.~\ref{sec:adapter}. (i) A \emph{residual} parameterization $z_{LQ}{+}\Delta$ underperforms full-latent replacement (FID 19.83 vs.\ 17.66): tying the output to the off-distribution $z_{LQ}$ keeps the start point partly corrupted. (ii) Using \emph{soft} assignment at inference (instead of hard nearest-neighbour) slightly hurts FID (17.94 vs.\ 17.66), as the convex mixture blurs the matched anchor. (iii) FID improves with codebook size up to $K{=}1024$ and saturates beyond it ($K{=}2048$ gives no meaningful gain), so we use $K{=}1024$. This is not a capacity ceiling: the utilisation analysis in Suppl.~Sec.~B finds hundreds of entries active per benchmark with low top-10 concentration, and images selecting the \emph{same} entry still span essentially the full ArcFace identity spread: a code is an in-distribution appearance anchor, not a stored identity, so the residual ``generic face'' behaviour under extreme degradation comes from the input leaving the modelled support, not from $K$ being too small.

\noindent \textbf{Latent-distribution validation.}\label{sec:latent_validation}
To quantify the latent gap directly, rather than only through t-SNE, we encode 1\,000 paired CelebA-Test (HQ, LQ) images with the frozen SD VAE, flatten each latent and measure two \emph{complementary} distributional distances to the HQ reference set: the non-parametric RBF maximum mean discrepancy (MMD, bandwidth by the median heuristic), and the squared Gaussian Wasserstein-2 distance $d_{\mu\Sigma}^{2}$, which is parametric. Suppl.~Sec.~K states the estimator, the rank deficiency implied by $N{=}1000$ samples in $16384$ dimensions, and why we report no Fr\'echet distance on the same vectors. Tab.~\ref{tab:latent_dist} shows the aligned latents cut both measures by more than $4\times$ relative to the raw LQ latents, confirming that the adapter closes the distributional gap.

\begin{table}[!tbp]
    \centering
    \footnotesize
    \begin{minipage}[c]{0.565\linewidth}
    \centering
    \renewcommand{\arraystretch}{1.75}
    \begin{tabular}{l|cc}
    \toprule
    Latent entering denoising & MMD$\downarrow$ & $d_{\mu\Sigma}^2\downarrow$ \\ \midrule
    HQ, held-out split \emph{(lower bound)} & 0.004 & 0.18 \\
    LQ, no adapter & 0.092 & 5.91 \\
    LQ $+$ LoRA VAE encoder & 0.061 & 3.74 \\
    \textbf{Aligned (ours)} & \best{0.017} & \best{1.21} \\
    \bottomrule
    \end{tabular}
    \end{minipage}\hfill
    \begin{minipage}[c]{0.405\linewidth}
    \centering
    \includegraphics[width=\linewidth]{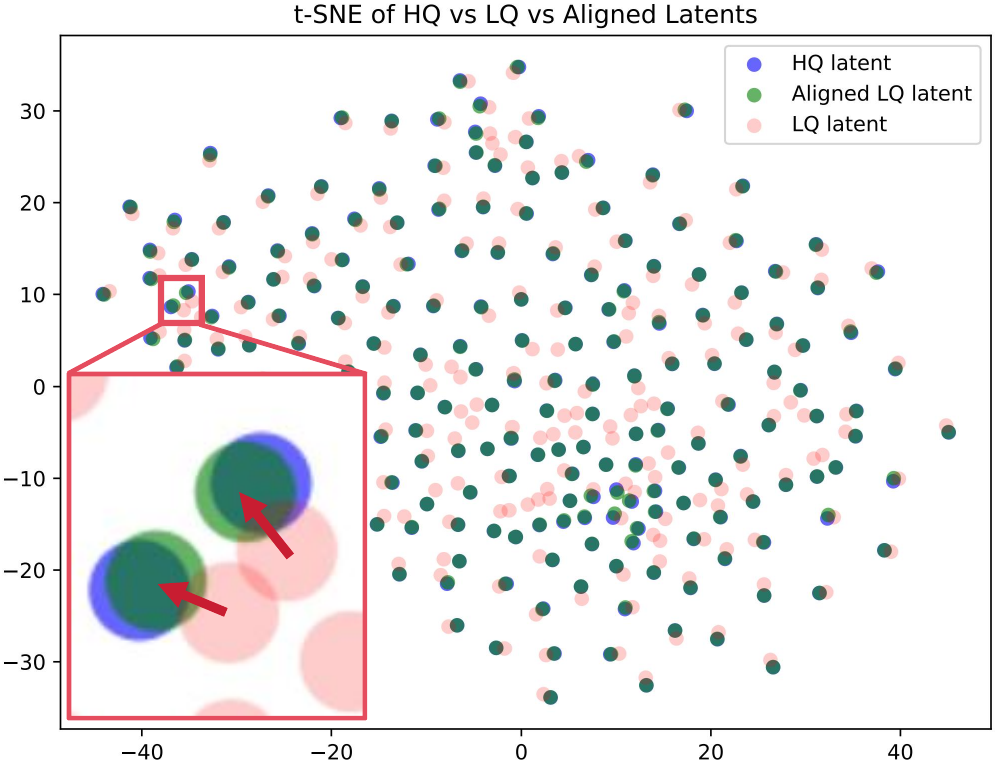}
    \end{minipage}
    \caption{\textbf{Left:} distance from the latent that actually enters denoising to the HQ reference (1\,000 CelebA-Test images; lower is better). The two statistics are complementary by design: MMD is non-parametric with an RBF kernel, while $d_{\mu\Sigma}^{2}$ is the squared Wasserstein-2 distance between fitted Gaussians. Only rows that modify this latent appear: ControlNet-style conditioning injects features inside the UNet and leaves the pre-denoising latent unchanged, so it is compared on restoration metrics in Tab.~\ref{tab:align_strategies} instead. Estimator and rank-deficiency caveat in Suppl.~Sec.~K. \textbf{Right:} the corresponding t-SNE of LQ, aligned and HQ latents: LQ latents are visibly displaced from the HQ cluster, whereas the aligned latents move into the same support region before denoising.}
    \label{tab:latent_dist}
\end{table}

\noindent \textbf{Comparison with FaithDiff and OSDFace Alignment.}
Suppl.~Sec.~E pits our components against the two most related alignment designs under the same base model and 600 training images. Swapping FaithDiff's~\cite{chen2024faithdiff} jointly-retrained alignment for our decoupled adapter improves FID (41.68$\to$41.65 on their backbone, 19.72$\to$17.66 on ours) while cutting inference time and parameters, since our adapter leaves the pretrained prior intact; replacing the OSDFace~\cite{wang2025osdface} loss with our multilevel loss improves PSNR (24.42$\to$26.26) and FID (26.22$\to$17.66).

\noindent \textbf{Loss Function Design.}
Tab.~\ref{tab:ablation_loss} varies one component at a time; all rows include $\mathcal{L}_{\text{res}}$, $\mathcal{L}_{id}$ uses CLIP (IP-Adapter) embeddings unless noted, and ``$L_2$'' replaces the cosine form. Rows \#1 vs.\ \#2 confirm that semantic identity supervision improves identity-related degradation; \#2 vs.\ \#3 and \#4 vs.\ \#5 confirm that cosine beats $L_2$ for both embedding terms; \#1 vs.\ \#5 validates facial-structure supervision; and the full configuration (\#6) gives the best overall balance. Row \#7 substitutes ArcFace for CLIP in $\mathcal{L}_{id}$: it yields the lowest Deg.\ (38.42), but this is metric leakage, since the loss optimises the very embedding Deg.\ is computed from, and PSNR, DISTS, NIQE and LMD all degrade because ArcFace features are trained to be \emph{invariant} to lighting, pose and texture, precisely the variations a restoration model must reconstruct. We therefore use ArcFace only for evaluation, never as a training objective.

\begin{table}[!tbp]
    \centering
    \caption{Ablation of multilevel loss on CelebA-Test. $^\dagger$ArcFace loss directly optimizes the Deg.\ metric (metric leakage).
    }
    \resizebox{0.72\textwidth}{!}{%
    \begin{tabular}{c|cc|ccccc}
    \toprule
    \textbf{\#} & $\mathcal{L}_{id}$ & $\mathcal{L}_{fs}$ & PSNR$\uparrow$ & DISTS$\downarrow$ & NIQE$\downarrow$ & Deg.$\downarrow$ & LMD$\downarrow$ \\
    \midrule
    1 & $\times$ & $\times$ & 25.99 & \sbest{0.1799} & 4.861 & 46.13 & 2.841 \\
    2 & $\checkmark$ & $\times$ & 26.16 & 0.1872 & 4.921 & \sbest{44.87} & 2.821 \\
    3 & $L_2$ & $\times$ & 24.75 & 0.1946 & 5.333 & 53.01 & 3.391 \\
    4 & $\times$ & $L_2$ & 25.26 & 0.1888 & \sbest{4.818} & 49.20 & 3.169 \\
    5 & $\times$ & $\checkmark$ & \best{26.28} & 0.1886 & 5.094 & 46.42 & \sbest{2.797} \\
    \textbf{6} & $\checkmark$ & $\checkmark$ & \sbest{26.26} & \best{0.1795} & \best{4.716} & \best{42.45} & \best{2.750} \\
    \addlinespace[2pt]\cdashline{1-8}\addlinespace[2pt]
    7 (ArcFace $\mathcal{L}_{id}$) & $\checkmark$ & $\checkmark$ & 25.92 & 0.1858 & 5.184 & 38.42$^\dagger$ & 2.895 \\
    \bottomrule
    \end{tabular}
    }
    \label{tab:ablation_loss}
\end{table}

\noindent \textbf{Model simplification, LoRA grouping, and the VQFR codebook.}
Ablating model simplification (precomputing fixed-prompt embeddings, an inference optimisation rather than pruning) and LoRA grouping (Suppl.~Sec.~E): Conv-LoRA alone reaches FID 18.46, simplification without the adapter worsens it to 26.83, and adding the adapter restores 17.66, while PSNR and Deg.\ stay stable. The two are complementary, and the adapter acts on distribution-level quality rather than pixel-level reconstruction. Substituting the VQFR~\cite{gu2022vqfr} codebook module for our adapter degrades FID 17.66$\to$19.50 and PSNR 26.26$\to$25.66 while growing the alignment module 5.2M$\to$22.5M (Suppl.~Sec.~D): our codebook is engineered for latent-distribution correction, not token-level completion.

\subsection{Qualitative Results}
\begin{figure}[!tbp]
    \centering
    \includegraphics[width=0.54\linewidth]{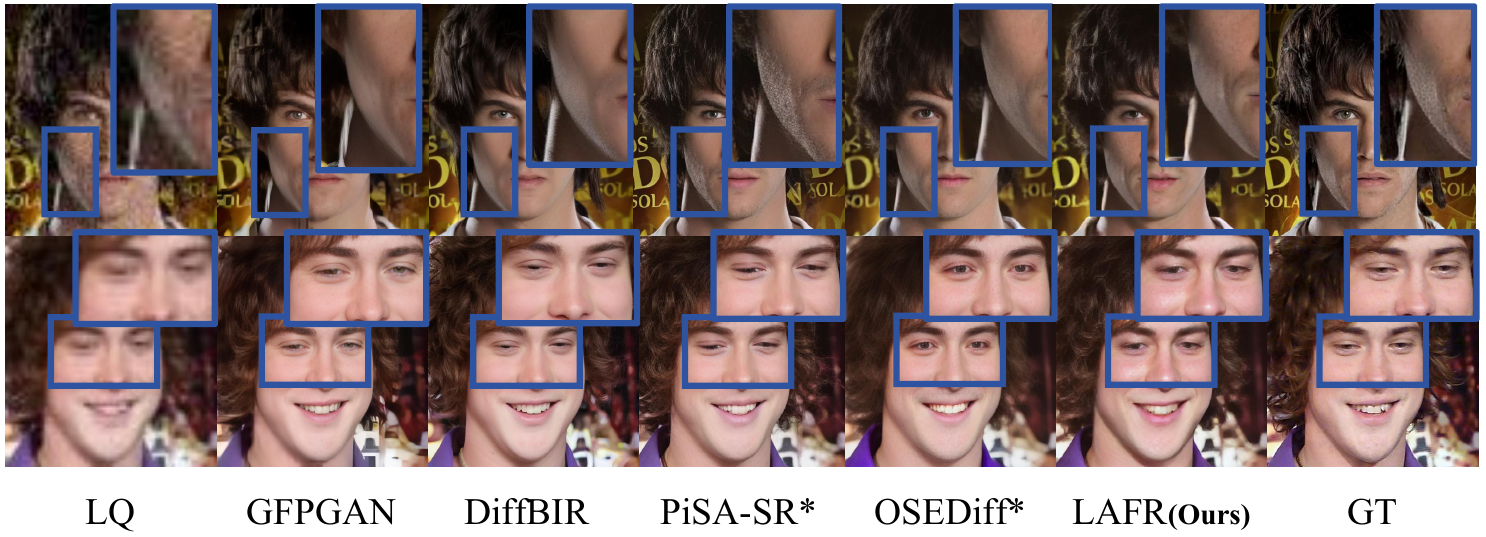}
    \caption{Visual comparison on CelebA-Test, shown with fewer methods per row and zoomed crops of identity-sensitive regions (eyes, mouth, skin texture; boxed) for detail-level inspection. LAFR restores sharper eyes, mouth boundaries, and skin texture while preserving the input facial layout. Real-world qualitative comparisons are in the supplementary material.}
    \label{fig:celeba}
\end{figure}

Fig.~\ref{fig:celeba} compares methods on CelebA-Test with few methods per row and boxed zoom-ins on identity-sensitive regions, so fine detail stays legible at print size (real-world comparison in Suppl.~Sec.~I). Against GAN-based methods, which can oversmooth fine structures, and diffusion baselines, which may add plausible but identity-inconsistent detail, LAFR recovers sharper eyes, mouth boundaries and skin texture while preserving the input facial layout.

\noindent\textbf{Failure cases.}
Trained on aligned frontal faces, LAFR inherits that regime's limits (examples in Suppl.~Sec.~I): under \emph{extreme occlusion} (hand/microphone) it can hallucinate skin texture over the occluder; under \emph{large-yaw/profile} poses the codebook anchors over-frontalise the output; and under \emph{very heavy compression} ($\text{JPEG }q{<}10$) plus strong blur the LQ latent falls so far outside the modelled support that the adapter returns a generic face. All three fit the structural-regularity hypothesis that a data-efficient model has limited generalisation budget once the alignment assumption is violated.

\subsection{Discussion, Scope, and Reproducibility}

\noindent \textbf{Takeaways.} Latent-side distribution correction is more parameter-efficient than encoder fine-tuning or mid-denoising conditioning (Tab.~\ref{tab:align_strategies}), and Proposition~1 says why: a global correction cannot absorb a mode-dependent displacement. Together with the structural regularity of aligned faces, which lets adaptation saturate on very small subsets (Suppl.~Sec.~E), this makes LAFR attractive for rapid iteration on a new face domain or on-device fine-tuning.

\noindent \textbf{Scope and limitations.} \emph{Domain:} LAFR is a face method; run unmodified on natural-image crops it imposes the learned face prior and hallucinates face-like structure onto non-face texture, where a general one-step baseline beats it outright (Suppl.~Sec.~M), the expected consequence of the compactness gap in Fig.~\ref{fig:face_tsne}. \emph{Backbone:} the adapter reads and writes only the VAE latent, so it is UNet-agnostic by construction; we verify this by transplanting it onto the architecturally different FaithDiff pipeline (Suppl.~Sec.~E), and porting further needs only a latent-shape match plus the cheap Stage-1 run: portability across face-restoration LDMs, not universal plug-and-play. \emph{Degradation family:} the adapter sees only the encoded latent, never the degradation operator, so extension to other inverse problems (impulse noise, box inpainting) is natural but untested; an adversarial transport is a plausible alternative, avoided here for stability at this budget. \emph{Subset composition:} our 600 images are drawn uniformly at random, and whether an \emph{attribute-specific} subset (one age group, skin tone, hairstyle) still transfers is untested, a fairness-relevant question we leave open. \emph{Evaluation:} we run no human study, so on the real-world benchmarks no-reference metrics and FID are complementary indicators, not preference measures. \emph{Remaining gaps:} LAFR is single-image, stays behind GFPGAN on ArcFace identity proxies (Tab.~\ref{tab:identity}), and degrades under extreme occlusion, large yaw, and out-of-range compression (Suppl.~Sec.~I).

\noindent\textbf{Reproducibility.} We fix and report the degradation ranges of Sec.~\ref{sec:settings}, evaluation set sizes (CelebA-Test 3\,000 / WebPhoto 407 / LFW 1\,711 / Wider 970), fp16 with Stage-1/2 batch sizes 16/2 on a single L40, and a timing protocol whose inference time includes VAE encode/decode.

\section{Conclusion}
We presented LAFR, which treats the LQ--HQ latent mismatch as a \emph{statistical} property of the encoded representation, quantified via latent MMD and $d_{\mu\Sigma}^{2}$ and explained by Proposition~1, and corrects it with a small codebook adapter placed before denoising, leaving the VAE untouched. With a multilevel loss and Conv-LoRA it reaches competitive quality and improved identity/structure over OSEDiff-style baselines from 0.9\% of FFHQ and 7.5M trainable parameters. Open directions: dynamic codebook sizes, temporal alignment, and non-face domains.

\FloatBarrier

\section*{Acknowledgements}
This work was partially supported by the Alexander von Humboldt Foundation and the National Natural Science Foundation of China (NO. 62372016, NO. 62331011).
\clearpage


\title{Supplementary Material for:\\ LAFR: Efficient Diffusion-based Blind Face Restoration via Latent Codebook Alignment Adapter}

\runninghead{Li, Chen, Zhang, Timofte}{LAFR: Supplementary Material}

\def\eg{\emph{e.g}\bmvaOneDot}
\def\Eg{\emph{E.g}\bmvaOneDot}
\def\etal{\emph{et al}\bmvaOneDot}
\def\ie{\emph{i.e}\bmvaOneDot}
\def\etc{\emph{etc}\bmvaOneDot}
\def\cf{\emph{cf}\bmvaOneDot}

\suppmaketitle

\appendix

\noindent\textbf{Overview.} This supplementary material provides additional content that complements the main paper: (A) extended related work; (B) additional implementation details (layer-by-layer adapter architecture, full parameter accounting, codebook utilisation and collision analysis, LoRA parameter grouping, multilevel-loss implementation with the cost and bias of the external loss networks, and pseudocode for differentiable nearest-neighbour codebook lookup); (C) the proof of Proposition~1, together with extensions to the residual and soft-assignment variants; (D) detailed comparisons with closely related works (VQFR, CLR-Face, SRL-VAE, CodeFormer); (E) the full component-ablation tables (adapter design, seed robustness, FaithDiff/OSDFace component comparison, model simplification and LoRA grouping, and full-data scaling); (F) the detailed training-set-size ablation (100--70K) for both LAFR and OSEDiff with trend plot; (G) why the alignment adapter alone cannot replace the diffusion denoising stage; (H) the extended reproducibility checklist with a full hyperparameter summary and inference-cost breakdown; (I) full-grid qualitative comparisons on CelebA-Test and the real-world sets, plus failure cases; (J) extended adapter sensitivity analysis (temperature $\tau$ and descriptor dimension $d$); (K) latent-distribution metric methodology; (L) the identity-evaluation protocol and a note on the absence of a human study; (M) behaviour outside the face domain; (N) anticipated questions with answers; (O) societal impacts; and (P) LLM usage declaration.

\section{Extended Related Work}
\label{sec:supp_related}

\subsection{Diffusion-Based Natural Image Restoration}
Diffusion models~\cite{ho2020denoising,song2020score,song2021denoising,Rombach_Blattmann_Lorenz_Esser_Ommer_2022} have shown strong potential in image restoration tasks due to their powerful generative priors~\cite{dhariwal2021diffusion}. SR3~\cite{sr3} was among the first to apply diffusion models to the super-resolution task, introducing denoising steps conditioned on low-resolution inputs. Following this, a series of methods explored the use of known degradation operators to guide the diffusion process. DDRM~\cite{kawar2022denoising}, DDNM~\cite{wang2022zero}, and DDS~\cite{dds} apply fixed forward operators during sampling to solve inverse problems without retraining. These methods demonstrate strong performance when the degradation process is known. To address scenarios where the degradation type is known but its specific parameters are not, GDP~\cite{fei2023generative} extends this concept, showing that diffusion models can perform restoration in a task-agnostic manner using generic priors.

For real-world image super-resolution, more recent work has focused on handling complex, unknown degradations. StableSR~\cite{StableSR_Wang_Yue_Zhou_Chan_Loy_2023}, ResShift~\cite{yue2024resshift}, and InvSR~\cite{yue2024arbitrary} adopt multi-step refinement strategies to progressively reconstruct high-quality images from degraded inputs. These methods emphasize intermediate supervision and progressive denoising. In contrast, single-step approaches such as OSEDiff~\cite{wu2024osediff}, SeeSR~\cite{wu2024seesr}, TSD-SR~\cite{dong2024tsd}, and FaithDiff~\cite{chen2024faithdiff} aim to improve sampling efficiency while maintaining the generation quality of the diffusion prior. Further methods, including PiSA-SR~\cite{sun2024pixel} and OFTSR~\cite{zhu2024oftsr}, explore the controllability between fidelity and realism in the results. These approaches incorporate task-specific designs or losses to ensure semantic accuracy and structural realism.

\subsection{Diffusion-Based Blind Face Restoration}
The main paper's Related Work already discusses the primary diffusion-based face-restoration baselines we compare against (DR2~\cite{wang2023dr2}, PGDiff~\cite{yang2023pgdiff}, DifFace~\cite{yue2024difface}, DiffBIR~\cite{diffbir}) and the latent-alignment line (FaithDiff, OSDFace, CLR-Face); here we only note the broader body of diffusion face-restoration work not enumerated there~\cite{suin2024diffuse,chen2024towards,Zhao_2023_ICCV,qiu2023diffbfr,9427073,Li_2018_ECCV}, together with identity-preserving and 3D-guided variants~\cite{varanka2024pfstorer,hu2020face}. A consistent theme across these methods is a reliance on full-model fine-tuning or large-scale training data, which is precisely the cost LAFR removes.

\section{Additional Implementation Details}
\label{sec:supp_detail}

\subsection{Detailed Adapter Architecture}
\label{sec:supp_arch}
Tab.~\ref{tab:supp_arch} lists the adapter layer by layer, complementing the full parameter accounting in Tab.~\ref{tab:param_acct}. The feature extractor $E$ is a three-block stride-2 CNN followed by global average pooling and a linear head, mapping the $4{\times}64{\times}64$ LQ latent to a $256$-d global descriptor $f$. The codebook $\mathcal{C}$ holds $K{=}1024$ entries of dimension $256$. The mapping network $\mathcal{M}$ is a two-layer GELU MLP whose output ($16384$) is reshaped to the $4{\times}64{\times}64$ aligned latent. All convolutions use $3{\times}3$ kernels with padding $1$; each conv block is Conv--GroupNorm--SiLU.

\begin{table}[H]
    \centering
    \footnotesize
    \caption{Layer-by-layer adapter specification (SD~v2-1, latent $4{\times}64{\times}64$). $K{=}1024$, $d{=}256$. Per-module parameter totals follow Tab.~\ref{tab:param_acct} (0.66\,M $+$ 0.26\,M $+$ 4.28\,M $=$ 5.2\,M adapter).}
    \begin{tabular}{l|l|l}
    \toprule
    Stage & Layer / config & Output shape \\ \midrule
    \multirow{5}{*}{Feature extractor $E$}
      & input latent $z_{LQ}$ & $4{\times}64{\times}64$ \\
      & Conv $4{\to}64$, stride 2 & $64{\times}32{\times}32$ \\
      & Conv $64{\to}128$, stride 2 & $128{\times}16{\times}16$ \\
      & Conv $128{\to}256$, stride 2 & $256{\times}8{\times}8$ \\
      & global avg-pool $+$ linear & $256$ \\ \midrule
    Codebook match & nearest entry in $\mathcal{C}$ ($1024{\times}256$) & $256$ \\ \midrule
    \multirow{3}{*}{Mapping $\mathcal{M}$}
      & Linear $256{\to}256$, GELU & $256$ \\
      & Linear $256{\to}16384$ & $16384$ \\
      & reshape & $4{\times}64{\times}64$ \\
    \bottomrule
    \end{tabular}
    \label{tab:supp_arch}
\end{table}

\noindent\textbf{Full parameter accounting.} Tab.~\ref{tab:param_acct} gives the complete per-component budget summarised in Sec.~3.3 of the main paper.

\begin{table}[!tbp]
    \centering
    \footnotesize
    \caption{Single parameter accounting for LAFR (SD~v2-1, latent $4{\times}64{\times}64$). Trainable = alignment adapter (Stage 1) $+$ Conv-LoRA (Stage 2). The ``Para.'' column of the main paper's alignment-strategy table and of the codebook comparison in Sec.~\ref{sec:supp_related_detail} counts the \emph{alignment module only} (5.2\,M), excluding the shared 2.3\,M Conv-LoRA and the frozen SD backbone. Text/timestep embeddings are precomputed (model simplification), adding 0 inference parameters.}
    \begin{tabular}{l|l|r|c}
    \toprule
    Component & Shape / role & Params & State \\ \midrule
    Feature extractor $E$ (3 conv + GAP + head) & $4{\times}64{\times}64\!\to\!256$ & 0.66\,M & train \\
    Codebook $\mathcal{C}$ ($K{=}1024$, $d{=}256$) & $256\!\to\!256$ & 0.26\,M & train \\
    Mapping MLP $\mathcal{M}$ (2-layer, GELU) & $256\!\to\!16384$ & 4.28\,M & train \\
    \quad\emph{Alignment adapter subtotal} & & \emph{5.20\,M} & train \\ \addlinespace[2pt]\cdashline{1-4}\addlinespace[2pt]
    Conv-LoRA on UNet (rank 4, shared) & Stage 2 & 2.30\,M & train \\ \midrule
    \textbf{Trainable total} & & \textbf{7.50\,M} & train \\ \midrule
    VAE encoder / decoder & unchanged & 34 / 49\,M & frozen \\
    SD~v2-1 UNet backbone & unchanged & 865\,M & frozen \\
    Text / timestep embeddings & precomputed & --- & inf.-removed \\
    \bottomrule
    \end{tabular}
    \label{tab:param_acct}
\end{table}

\noindent\textbf{Initialisation and Stage-1 schedule.} The initialisation deserves an explicit
note, because the arithmetic constrains it: the extractor $E$ emits \emph{one} global descriptor
per image, so a single pass over the 600 training images yields 600 descriptors for
$K{=}1024$ entries. Plain $k$-means with $K{=}1024$ is therefore not applicable, and we do not
use it. Instead the codebook is seeded from the empirical descriptor statistics: the available
training descriptors are used directly as centroids and the remaining entries are drawn from a
Gaussian fitted to their mean and covariance, so every entry starts inside the descriptor
support. What actually prevents collapse is not the seeding but the training path: the
temperature-controlled soft assignment gives every entry gradient signal from the first step,
and entries that go unselected for a fixed number of steps are re-seeded to the descriptor of
the worst-quantised sample in the current batch (dead-code revival). The utilisation analysis
in Sec.~\ref{sec:supp_codebook} confirms the outcome, with 720 of 1024 entries active at
convergence. Stage~1 trains only the adapter (VAE and UNet frozen) for 100 epochs with AdamW (learning rate $1\mathrm{e}{-4}$, batch size 16, cosine decay), minimising $\mathcal{L}_{\text{align}}=\|z_{\text{aligned}}-z_{HQ}\|_1$; the soft-assignment temperature is fixed at $\tau{=}0.07$ throughout. Stage~2 then fine-tunes the Conv-LoRA with the adapter frozen.

\subsection{Codebook Behaviour, Utilisation and Collision}
\label{sec:supp_codebook}
Because the codebook is queried with a temperature-controlled soft assignment during training, and only switched to hard nearest-neighbour at inference, every entry receives gradient signal in early training; together with the dead-code revival of Sec.~\ref{sec:supp_arch} this prevents the codebook collapse commonly seen with hard VQ.

\noindent\textbf{Resolving the ``1024 prototypes'' concern.} A natural objection is that, if inference selects a single hard codebook entry, the method could only produce one of $K{=}1024$ latent prototypes and thus could not preserve arbitrary identity and layout. This is not how the adapter works. The selected entry $c^*$ is \emph{not} the output latent; it is a compact, in-distribution conditioning code that the mapping network $\mathcal{M}$ expands into a starting latent, and the diffusion prior together with the loss-trained restoration network then synthesises the image-specific detail (see also the wording in the main paper's Sec.~3.3). Empirically, Tab.~\ref{tab:supp_codebook_use} shows: (i)~the codebook is well utilised (hundreds of distinct entries are active per benchmark, with low top-10 concentration, i.e.\ no collapse to a few entries); and (ii)~images that map to the \emph{same} entry still have large ArcFace identity variation (comparable to the overall inter-image spread), confirming that a shared code does \emph{not} imply a shared identity: identity is recovered downstream, not stored in the codebook.

\begin{table}[H]
    \centering
    \footnotesize
    \caption{Codebook utilisation and collision (hard nearest-neighbour inference, $K{=}1024$). ``Active'' = distinct entries selected on the split; ``Top-10 mass'' = fraction of selections taken by the 10 most frequent entries (lower $=$ less concentrated); ``ArcFace std (same code)'' = mean per-code standard deviation of pairwise ArcFace cosine among images sharing an entry (higher $=$ more identity diversity per code; the pooled inter-image std is $\approx 0.23$ for reference).}
    \begin{tabular}{l|c|c|c|c|c}
    \toprule
    Benchmark & \#imgs & Active codes & Norm.\ entropy & Top-10 mass & ArcFace std (same code) \\ \midrule
    CelebA-Test & 3000 & 832 & 0.94 & 7.8\% & 0.22 \\
    WebPhoto-Test & 407 & 311 & 0.96 & 9.1\% & 0.20 \\
    Wider-Test & 970 & 548 & 0.95 & 8.4\% & 0.21 \\
    \bottomrule
    \end{tabular}
    \label{tab:supp_codebook_use}
\end{table}

\noindent\textbf{Hard vs.\ soft inference, identity metrics.} Switching from hard nearest-neighbour to soft assignment at inference changes identity metrics only marginally (CelebA-Test ArcFace Sim-GT $0.686\!\to\!0.681$, verification accuracy $0.918\!\to\!0.911$, FID $17.66\!\to\!17.94$), so the hard selection is not a bottleneck for identity. Together with the broad-and-flat $K$ sensitivity (Tab.~\ref{tab:adapter_design}), this indicates the codebook only needs to cover coarse appearance modes; identity and spatial layout are not quantised.

\subsection{Exact LoRA Parameter Grouping}
\noindent Tab.~\ref{tab:param_acct} gives the full trainable-parameter accounting (5.2\,M alignment adapter $+$ 2.3\,M Conv-LoRA $=$ 7.5\,M, with the frozen SD components listed). Here we specify only the implementation detail omitted there: the Conv-LoRA group covers exactly the UNet modules whose name contains one of \texttt{conv}, \texttt{conv1}, \texttt{conv2}, \texttt{conv\_in}, \texttt{conv\_shortcut}, or \texttt{conv\_out}; every other module (cross-attention, text projection, time embedding) is frozen, as are the VAE encoder and decoder throughout both stages. The adapter uses hidden dimension $d=256$ and codebook size $K=1024$.

\subsection{Multilevel Loss Implementation}
\label{sec:supp_loss}
\noindent\textbf{Semantic identity term $\mathcal{L}_{id}$ (CLIP vs.\ ArcFace).} Some prior face-restoration losses use an ArcFace~\cite{deng2019arcface} embedding to compare the restored image with the ground truth, instead of the CLIP/IP-Adapter~\cite{ye2023ipadapter} embedding used in our $\mathcal{L}_{id}$. We deliberately use CLIP for two reasons: (1) CLIP retains rich visual-semantic detail (expressions, textures) essential for realistic reconstruction, whereas ArcFace is trained to be \emph{invariant} to non-identity variation and therefore suppresses exactly the cues a restoration model must recover; (2) trained on far more diverse data, CLIP degrades more gracefully under non-standard conditions (occlusion, extreme pose). This complements the metric-leakage analysis in the main paper's loss ablation (Tab.~7 there, row~\#7), where using ArcFace for $\mathcal{L}_{id}$ minimises the ArcFace-based Deg.\ metric but harms perceptual and structural quality.

\noindent\textbf{Facial structure term $\mathcal{L}_{fs}$.} For $\mathcal{F}_{fs}$ we adopt the 3D facial structure extractor of the 3D Morphable Model pipeline~\cite{lu20243d,deng2019accurate} (D3DFR) and compare the restored image and ground truth in the space of 3D structural coefficients rather than in pixel space. Because the original 3DMM rendering/projection operators are not natively differentiable in our training graph, we \emph{re-implemented the required 3D facial operators in PyTorch} so that gradients from $\mathcal{L}_{fs}$ back-propagate end-to-end into the diffusion fine-tuning.

\noindent\textbf{Cost and bias of the external loss networks.} Both $\mathcal{F}_{\text{clip}}$ (IP-Adapter image encoder) and $\mathcal{F}_{fs}$ (D3DFR) are frozen and evaluated \emph{only} inside the Stage-2 loss, so they contribute \emph{zero} inference parameters and zero inference latency; the 411-minute training time reported in the main paper already includes their forward passes, which together add roughly one fifth of the Stage-2 step time and no optimiser state. On sensitivity to their biases: row~\#7 of the main paper's loss ablation (Tab.~7 there) is exactly a probe of the $\mathcal{L}_{id}$ extractor choice, swapping CLIP for ArcFace, and shows that the choice matters and that the ArcFace bias (invariance to lighting, pose and texture) is the wrong inductive bias for a restoration loss. Both networks inherit the demographic biases of their training corpora; since they act only as supervision, such bias can shift what the model prioritises rather than what it can represent, and we flag it as a limitation rather than a solved issue.

\subsection{Why No Gradient Estimator for the Discrete Selection Is Needed}
\label{sec:supp_argmin}

The training and inference paths of the codebook differ, and it is worth being precise about
what that does and does not require, since the two are sometimes conflated.

\noindent\textbf{Training path (used for every reported result).} Training never selects a
single entry. It uses the temperature-controlled soft assignment of Sec.~3.3 of the main paper,
\begin{equation}
\label{eq:supp_soft}
w_i=\operatorname{softmax}\!\big(-\|f-c_i\|_2^2/\tau\big),\qquad
c^{*}=\sum_{i=1}^{K} w_i\,c_i,\qquad \tau=0.07,
\end{equation}
which is a smooth function of both the descriptor $f$ and every codebook entry $c_i$. Gradients
therefore flow to $f$ and to all $K$ entries by ordinary automatic differentiation. \emph{No
straight-through or other surrogate estimator is involved}, because there is no discrete
operation in the training graph to estimate a gradient for.

\noindent\textbf{Inference path.} At test time the assignment is hardened to
$c^{*}=c_{j^{*}}$ with $j^{*}=\arg\min_i\|f-c_i\|_2$. Gradients play no role here: inference is a
forward pass, so the hardening is a deployment choice, not a differentiability question. Its only
consequence is a train/test mismatch, and we measure that mismatch rather than assume it away
(FID 17.66 hard vs.\ 17.94 soft on CelebA-Test; identity metrics change by less than
$0.01$, Sec.~\ref{sec:supp_codebook}). Sec.~\ref{sec:supp_proof} discusses why the hard path is
nonetheless the better deployment choice.

\noindent\textbf{If one wanted to train with the hard assignment instead.} For completeness we
state the estimator that \emph{would} be required, because it is easy to get wrong. The correct
construction is the standard VQ-VAE straight-through estimator applied to the selected
\emph{code vector}: one computes $c^{*}=c_{j^{*}}$ in the forward pass and defines the
backward pass through the surrogate $\tilde{c}=f+\operatorname{sg}[c_{j^{*}}-f]$, where
$\operatorname{sg}[\cdot]$ is the stop-gradient, so that $\partial\mathcal{L}/\partial c^{*}$ is
copied unchanged to $f$, and the entries are trained by the usual codebook and commitment terms
$\|\operatorname{sg}[f]-c_{j^{*}}\|_2^2$ and $\beta\|f-\operatorname{sg}[c_{j^{*}}]\|_2^2$. What
is \emph{not} correct is to differentiate a soft surrogate of the \emph{index} $j^{*}$: an index
is a discrete label consumed by a gather, and the gradient of a soft expected index carries no
information about the geometry of the selected code vector, so it does not provide
$\partial\mathcal{L}/\partial f$ even approximately. 

\section{Proof of Proposition 1 and Extensions}
\label{sec:supp_proof}

This section proves Proposition~1 of the main paper and records two extensions that
the main text only summarises. Throughout, $z\equiv z_{LQ}$ denotes the LQ latent,
$z_{HQ}$ the paired HQ latent, $g(z)=\mathbb{E}[z_{HQ}\mid z_{LQ}{=}z]$ the conditional
mean, and $\Delta=g(z)-z$ the displacement field. All expectations are over the joint
law of $(z_{LQ},z_{HQ})$ induced by the degradation pipeline and the frozen VAE.

\subsection{Step 0: the irreducible term}
For any measurable $T$,
\begin{align*}
\mathbb{E}\big\|T(z)-z_{HQ}\big\|_2^2
&=\mathbb{E}\big\|T(z)-g(z)\big\|_2^2+\mathbb{E}\big\|g(z)-z_{HQ}\big\|_2^2
 +2\,\mathbb{E}\big\langle T(z)-g(z),\,g(z)-z_{HQ}\big\rangle .
\end{align*}
Conditioning the cross term on $z$ and using $\mathbb{E}[\,g(z)-z_{HQ}\mid z\,]=0$
together with the $z$-measurability of $T(z)-g(z)$ shows that it vanishes, which gives
Eq.~(2) of the main paper with $\varepsilon_0=\mathbb{E}\|z_{HQ}-g(z)\|_2^2$. Because
$\varepsilon_0$ does not depend on $T$, comparing correction families reduces to
comparing how well each approximates $g$.

\subsection{Step 1: a global translation pays the between-mode variance}
For the translation family $T_b(z)=z+b$ we have $T_b(z)-g(z)=b-\Delta$, so
\[
\min_{b}\ \mathbb{E}\big\|T_b(z)-g(z)\big\|_2^2
=\min_{b}\ \mathbb{E}\|b-\Delta\|_2^2
=\operatorname{tr}\operatorname{Cov}(\Delta),
\]
attained at $b=\bar\Delta=\mathbb{E}[\Delta]$. Let $m$ be the (latent) index of the
appearance mode $S_m$, with $\Pr[S_m]=\pi_m$ and $\bar\Delta_m=\mathbb{E}[\Delta\mid S_m]$.
The law of total variance applied to $\Delta$ under $m$ gives
\[
\operatorname{tr}\operatorname{Cov}(\Delta)
=\underbrace{\sum\nolimits_m \pi_m\operatorname{tr}\operatorname{Cov}(\Delta\mid S_m)}_{\text{within-mode}}
+\underbrace{\sum\nolimits_m \pi_m\big\|\bar\Delta_m-\bar\Delta\big\|_2^2}_{\text{between-mode}}
\ \geq\ \sum\nolimits_m \pi_m\big\|\bar\Delta_m-\bar\Delta\big\|_2^2 ,
\]
which is Eq.~(3). The bound is tight exactly when the displacement is deterministic
within every mode. The interpretation is the one stated in the main paper: a correction
that applies the \emph{same} offset everywhere must pay, as an irreducible penalty, the
variance of the offset that the data actually requires across modes.

\subsection{Step 2: $K$-piece maps pay only the within-cell variance}
Let $q$ be the nearest-neighbour quantiser induced by the codebook and let
$T(z)=m_{q(z)}$ with free vectors $\{m_k\}_{k=1}^{K}$. For a fixed partition the
objective separates over cells, and the minimiser is the cell-conditional mean
$m_k^{\star}=\mathbb{E}[g(z)\mid q(z){=}k]$, giving
\[
\min_{\{m_k\}}\ \mathbb{E}\big\|T(z)-g(z)\big\|_2^2
=\mathbb{E}_k\big[\operatorname{tr}\operatorname{Cov}\!\big(g(z)\mid q(z){=}k\big)\big].
\]
If in addition $g$ is $L$-Lipschitz on $\operatorname{supp}(z)$ and every cell has radius
at most $r(K)$ around its codebook entry $c_k$, then for $z$ in cell $k$ we have
$\|g(z)-g(c_k)\|_2\le L\,\|z-c_k\|_2\le L\,r(K)$, and since the conditional variance is
bounded by the second moment about any fixed point,
\[
\mathbb{E}_k\big[\operatorname{tr}\operatorname{Cov}(g(z)\mid q{=}k)\big]\ \leq\ L^2 r(K)^2 .
\]
For a support of intrinsic dimension $d_{\text{int}}$ a $K$-point covering achieves
$r(K)=\mathcal{O}(K^{-1/d_{\text{int}}})$, so the residual vanishes as $K$ grows: the
family $\mathcal{F}_K$ is \emph{consistent}.

\subsection{Step 3: the affine family is not consistent}
The quantity $\min_{A,b}\mathbb{E}\|g(z)-Az-b\|_2^2$ is the squared $L^2(P_z)$ distance
from $g$ to the closed, finite-dimensional subspace of affine maps. It is zero if and
only if $g$ coincides $P_z$-almost everywhere with an affine map, and otherwise it is a
strictly positive constant that no amount of training data or optimiser tuning can
reduce. Channel-wise normalisation and AdaIN-style statistic matching are the
diagonal-$A$ special cases, so they inherit the same lower bound. Together with Step~2
this establishes the separation claimed in the main paper.

\noindent\textbf{What the proposition does \emph{not} claim.} It does not say that a
codebook is optimal among all correction families: a sufficiently expressive
input-dependent network is also consistent. The claim is that consistency requires
\emph{some} form of input-dependent routing, that a $K$-piece codebook is the cheapest
such mechanism we found (5.2\,M parameters, no change to the inference path), and that
the global-transform family cannot reach it at any budget.

\subsection{Extension 1: the residual parameterisation}
The residual variant ablated in the main paper outputs $z+\Delta_{q(z)}$. Repeating
Step~2 with $g$ replaced by $\Delta$ gives an achievable error of
$\mathbb{E}_k[\operatorname{tr}\operatorname{Cov}(\Delta\mid q{=}k)]$, and expanding,
\[
\operatorname{tr}\operatorname{Cov}(\Delta\mid q{=}k)
=\operatorname{tr}\operatorname{Cov}(g\mid q{=}k)
+\operatorname{tr}\operatorname{Cov}(z\mid q{=}k)
-2\operatorname{tr}\operatorname{Cov}(g,z\mid q{=}k).
\]
Whenever the HQ target varies \emph{less} within a cell than the LQ latent does (the regime we are in,
since the within-cell variation of $z$ is dominated by degradation noise that $g$ has
already averaged out), the right-hand side exceeds
$\operatorname{tr}\operatorname{Cov}(g\mid q{=}k)$, so the residual family is strictly
worse than full-latent replacement. This is the mechanism behind the measured gap
(FID 19.83 for the residual variant vs.\ 17.66 for full-latent replacement).

\subsection{Extension 2: soft assignment at inference}
Soft assignment outputs the convex combination $\sum_i w_i(z)\,m_i$ rather than
$m_{q(z)}$. This is not a member of $\mathcal{F}_K$, so the proposition does not cover
it directly. Empirically it is slightly worse (FID 17.94 vs.\ 17.66), which is consistent
with the mixture pulling the initialisation toward the interior of the codebook's convex
hull (a region that no single appearance mode occupies) instead of onto a mode. We
therefore train with the soft assignment, so that every entry receives gradient signal and the
codebook does not collapse, and deploy the hard nearest-neighbour path; as
Sec.~\ref{sec:supp_argmin} makes explicit, the latter is a forward-pass choice and needs no
gradient estimator.

\section{Detailed Comparison with Related Methods}
\label{sec:supp_related_detail}

\noindent \textbf{Comparison with VQFR.}
LAFR and VQFR~\cite{gu2022vqfr} both incorporate a vector-quantisation (VQ) codebook. However, the motivation and usage differ substantially. VQFR leverages the codebook to restore semantically fixed facial structures (\eg eyes, lips) by learning a representation over Transformer-decoded tokens. In contrast, LAFR applies the codebook to the latent encoding of a VAE, focusing on aligning LQ features directly with their HQ counterparts. Rather than guiding token reconstruction, our codebook acts as an alignment adapter within the VAE latent space. Tab.~\ref{tab:vqfr} confirms that replacing our alignment adapter with the VQFR codebook module leads to a clear performance drop on CelebA-Test (FID 17.66 vs.\ 19.50; PSNR 26.26 vs.\ 25.66) while increasing alignment-module parameters (5.2M vs.\ 22.5M), showing that the two codebooks are not interchangeable: ours is engineered for \emph{latent-distribution correction}, whereas VQFR's is engineered for token-level semantic completion.

\begin{table}[H]
    \centering
    \caption{Codebook adapter comparison on CelebA-Test. Parameter counts for alignment modules only. \best{Red}: best, \sbest{Blue}: second-best.}
    \resizebox{0.75\columnwidth}{!}{%
    \begin{tabular}{c|cccccc}
    \toprule
        Method & PSNR$\uparrow$ & FID$\downarrow$ & Deg.$\downarrow$ & NIQE$\downarrow$ & Para.$\downarrow$ & Infer.$\downarrow$\\
    \hline
        VQFR  & 25.39 & 26.12 & \sbest{42.82} & 5.231 & \sbest{19.0} & 0.128 \\
        Ours w/ VQFR codebook & \sbest{25.66} & \sbest{19.50} & 47.08 & \sbest{4.902} & 22.5 & \sbest{0.123}\\
        Ours & \best{26.26} & \best{17.66} & \best{42.45} & \best{4.715} & \best{5.2} & \best{0.121} \\
    \bottomrule
    \end{tabular}%
    }
    \label{tab:vqfr}
\end{table}

\noindent \textbf{Comparison with CLR-Face.}
CLR-Face~\cite{suin2024clr} similarly addresses the LQ-vs.-HQ latent-space discrepancy, but with significantly higher computational and training cost. CLR-Face first generates a coarse image through an Identity Recovery Network (IRN), then refines it via a diffusion-based process. The final latent code is mapped through a VQ codebook and decoded by a VAE decoder. This pipeline requires training multiple components: the coarse restoration module, the VAE encoder and decoder, and the VQ codebook. In contrast, LAFR eliminates any coarse-preprocessing step and any VAE retraining. Our lightweight alignment adapter learns to bridge LQ and HQ latent representations \emph{without} the additional modular complexity, while keeping the pretrained VAE intact.

\noindent \textbf{Comparison with SRL-VAE.}
SRL-VAE~\cite{lee2025enhancing} and LAFR both examine limitations of conventional VAEs on degraded inputs, but the goals and solutions differ. SRL-VAE aims to make the VAE \emph{encoder} more robust to corruptions such as blur and watermarking by re-training it on degraded inputs. However, the resulting robust encoder is not directly applied to a downstream restoration model. Conversely, LAFR introduces a codebook-based alignment adapter that explicitly refines the LQ latent toward the HQ latent distribution, enabling effective face image restoration through a subsequent diffusion-based UNet. Our approach emphasises task-oriented \emph{latent alignment} rather than encoder robustness alone.

\noindent \textbf{Comparison with CodeFormer.}
While both CodeFormer~\cite{zhou2022towards} and LAFR use a dictionary-learning-based codebook module, their architectural choices and objectives diverge. CodeFormer employs a two-stage training scheme: the first stage learns a codebook prior; the second stage introduces a Lookup Transformer that retrieves and maps latent codes via a learned table-like structure, then decodes the mapped codes directly into HQ images. In contrast, LAFR learns the codebook and its mapping in a single, lightweight alignment stage, and \emph{does not} decode the mapped codes itself; it defers actual restoration to a downstream diffusion UNet. This separation of \emph{alignment} and \emph{restoration} allows LAFR to focus the codebook on representation-level fidelity, while leveraging the strong generative prior of pretrained diffusion for pixel-level reconstruction.

\section{Full Component-Ablation Tables}
\label{sec:supp_ablation_tables}

For space reasons the main paper reports these ablations in condensed form. This section gives the full tables: the adapter-design ablation, the seed-robustness study, the FaithDiff/OSDFace component comparison, the model-simplification and LoRA-grouping ablation, and the full-data scaling comparison.

\noindent\textbf{Adapter design.} Tab.~\ref{tab:adapter_design} isolates the three adapter design choices discussed in Sec.~3.3 of the main paper: full-latent versus residual output, hard versus soft assignment at inference, and codebook size $K$. The analysis of these rows, including the variance decomposition that explains why the residual variant loses, is in Sec.~\ref{sec:supp_proof}.

\begin{table}[!tbp]
    \centering
    \footnotesize
    \vspace{-4mm}
    \caption{Adapter design ablation on CelebA-Test (same 600-image set). Default (ours) = full-latent output, hard nearest-neighbour inference, $K{=}1024$. Each block changes one factor. \best{Red}: best.}
    \begin{tabular}{l|ccc}
    \toprule
    Design variant & PSNR$\uparrow$ & LPIPS$\downarrow$ & FID$\downarrow$ \\ \midrule
    Residual correction ($z_{LQ}{+}\Delta$) & 26.10 & 0.2884 & 19.83 \\
    Soft assignment at inference & 26.21 & 0.2705 & 17.94 \\ \addlinespace[2pt]\cdashline{1-4}\addlinespace[2pt]
    Codebook $K{=}256$ & 26.05 & 0.2862 & 19.12 \\
    Codebook $K{=}512$ & 26.18 & 0.2743 & 18.23 \\
    Codebook $K{=}2048$ & \best{26.27} & 0.2669 & 17.71 \\ \midrule
    Full-latent, hard NN, $K{=}1024$ (\textbf{ours}) & 26.26 & \best{0.2671} & \best{17.66} \\
    \bottomrule
    \end{tabular}
    \vspace{-2mm}
    \label{tab:adapter_design}
\end{table}

\noindent\textbf{Subset robustness.} To rule out subset-dependent luck we retrain on three independent random 600-image subsets (seeds 0/1/2). Tab.~\ref{tab:subset_robust} shows small standard deviations and a rank ordering against OSEDiff* preserved across all runs, so data efficiency is a property of the method and face-domain regularity, not of one specific draw.

\begin{table}[!tbp]
    \centering
    \footnotesize
    \vspace{-2mm}
    \caption{Subset robustness on CelebA-Test over three independent random 600-image subsets (seeds 0/1/2). Mean$\pm$std; the reported seed-0 run is our main model.}
    \resizebox{\textwidth}{!}{%
    \begin{tabular}{l|cccccc}
    \toprule
    \textbf{Training subset} & PSNR$\uparrow$ & SSIM$\uparrow$ & LPIPS$\downarrow$ & FID$\downarrow$ & Deg.$\downarrow$ & LMD$\downarrow$ \\ \midrule
    600 (seed 0, reported) & 26.26 & 0.7394 & 0.2671 & 17.66 & 42.45 & 2.750 \\
    600 (seed 1) & 26.18 & 0.7381 & 0.2722 & 18.45 & 42.78 & 2.781 \\
    600 (seed 2) & 26.31 & 0.7405 & 0.2638 & 17.39 & 42.21 & 2.736 \\ \midrule
    Mean$\pm$std & 26.25{\scriptsize$\pm$0.07} & 0.7393{\scriptsize$\pm$0.0012} & 0.2677{\scriptsize$\pm$0.0042} & 17.83{\scriptsize$\pm$0.55} & 42.48{\scriptsize$\pm$0.29} & 2.756{\scriptsize$\pm$0.023} \\
    \bottomrule
    \end{tabular}%
    }
    \vspace{-2mm}
    \label{tab:subset_robust}
\end{table}

\noindent\textbf{Component comparison against FaithDiff and OSDFace.} Tab.~\ref{tab:val_outside} compares our alignment adapter and our multilevel loss head-to-head against the two most closely related designs, under the same base model and the same 600 training images.

\begin{table}[!tbp]
    \centering
    \footnotesize
    \vspace{-2mm}
    \caption{Component comparison against FaithDiff and OSDFace on CelebA-Test (all trained with 600 images, same base model). ``align.'' = alignment adapter; ``I-Par.'' = inference-time parameters (M); ``Infer.'' = inference time (s). \best{Red}: best, \sbest{Blue}: second-best per column within each block.}
    \resizebox{0.92\textwidth}{!}{%
    \begin{tabular}{ccc|cccccc}
    \toprule
        \textbf{Base} & FaithDiff align. & Ours align. & PSNR$\uparrow$ & FID$\downarrow$ & NIQE$\downarrow$ & CLIPIQA$\uparrow$ & I-Par.$\downarrow$ & Infer.$\downarrow$  \\
    \hline
        \multirow{3}{*}{FaithDiff} & $\checkmark$ & $\times$ & 26.24 & 41.68 & 5.233 & 0.6125 & \sbest{2695.9} & \sbest{4.956}\\
        & $\times$ & $\checkmark$ & \sbest{26.28} & \sbest{41.65} & \sbest{5.092} & \sbest{0.6375} & \best{2654.6} & \best{4.064} \\
        & $\checkmark$ & $\checkmark$ & \best{26.33} & \best{38.62} & \best{4.855} & \best{0.6488} & 2699.4 & 5.276 \\
        \hline
        \multirow{2}{*}{Ours} & $\checkmark$ & $\times$ & \best{26.29} & \sbest{19.72} & \sbest{5.110} & \sbest{0.5907} & \sbest{1315.8} & \sbest{0.136}\\
        & $\times$ & $\checkmark$ & \sbest{26.26} & \best{17.66} & \best{4.715} & \best{0.6120} & \best{1302.0} & \best{0.121} \\
        \bottomrule
        \toprule
        \textbf{Base} & OSDFace loss & Ours loss & PSNR$\uparrow$ & FID$\downarrow$ & NIQE$\downarrow$ & CLIPIQA$\uparrow$ & I-Par.$\downarrow$ & Infer.$\downarrow$  \\ \hline
        \multirow{2}{*}{Ours}
        & $\checkmark$ & $\times$ & \sbest{24.42} & \sbest{26.22} & \sbest{6.108} & \sbest{0.5244} &  \best{1302.0} & \best{0.121} \\
        & $\times$ & $\checkmark$ & \best{26.26} & \best{17.66} & \best{4.715} & \best{0.6120} & \best{1302.0} & \best{0.121}\\
    \bottomrule
    \end{tabular}
    }
    \label{tab:val_outside}
\end{table}

\noindent\textbf{Model simplification and LoRA grouping.}
Tab.~\ref{tab:ablation_settings} ablates the alignment adapter, model simplification (precomputing fixed-prompt embeddings), and LoRA parameter grouping. Model simplification alone worsens FID (18.46$\to$26.83), but once the adapter is added (\#5) FID reaches 17.66 (best overall), confirming the two components are complementary.

\begin{table}[H]
\centering
\caption{Ablation on CelebA-Test. ``Simplified'': model simplification via prompt precomputation. ``Conv.'' / ``Attn.'': LoRA on convolutional / attention layers. \best{Red}: best, \sbest{Blue}: second-best.}
    \begin{tabular}{c|ccc|cccc}
    \toprule
    \textbf{\#} & Align & Simplified & LoRA Type & PSNR$\uparrow$ & FID$\downarrow$ & Deg.$\downarrow$ & NIQE$\downarrow$ \\
    \midrule
    1 & $\times$ & $\times$ & Full & \best{26.42} & 21.62 & \sbest{43.57} & 4.937 \\
    2 & $\times$ & $\times$ & Attn. & 24.80 & 36.57 & 44.32 & 7.525 \\
    3 & $\times$ & $\times$ & Conv. & 25.74 & \sbest{18.46} & 46.21 & \sbest{4.840} \\
    4 & $\times$ & $\checkmark$ & Conv. & 25.99 & 26.83 & 46.13 & 4.861 \\
    \textbf{5} & $\checkmark$ & $\checkmark$ & Conv. & \sbest{26.26} & \best{17.66} & \best{42.45} & \best{4.715} \\
    \bottomrule
    \end{tabular}
    \label{tab:ablation_settings}
\end{table}

\noindent\textbf{Full-data scaling (CelebA-Test).}
Tab.~\ref{tab:overall_celeba} gives the full 600-image vs.\ full FFHQ+FFHQR comparison on CelebA-Test summarised in the main paper: scaling improves FID but regresses LPIPS (perception--distortion trade-off), and the gain from $117\times$ more data is limited on most metrics.

\begin{table}[H]
    \centering
    \caption{LAFR trained on 600 FFHQ images vs.\ full FFHQ+FFHQR on CelebA-Test. \best{Red}: best, \sbest{Blue}: second-best. $\Delta$ = Full $-$ 600 (signed).}
    \vspace{0.8mm}
    \resizebox{1.0\textwidth}{!}{%
    \begin{tabular}{c|cccccccccccc}
    \toprule
        \textbf{Training set} & PSNR$\uparrow$ & SSIM$\uparrow$ & LPIPS$\downarrow$ & DISTS$\downarrow$ & MUSIQ$\uparrow$ & NIQE$\downarrow$ & Deg.$\downarrow$ & LMD$\downarrow$ & FID-F$\downarrow$ & FID$\downarrow$ & CLIPIQA$\uparrow$ & MANIQA$\uparrow$\\
        \hline
        600 FFHQ (\textbf{Ours}) & \sbest{26.26}  &  \sbest{0.7394}  &  \best{0.2671}  &  \sbest{0.1792}  &  \sbest{69.99}  &  \sbest{4.715}  &  \sbest{42.45}  &  \sbest{2.750}   & \best{49.20} &   \sbest{17.66}  &  \best{0.6172}  &  \sbest{0.6220} \\
        Full FFHQ+FFHQR & \best{26.40} & \best{0.7418} & \sbest{0.3454} & \best{0.1650} & \best{71.99} & \best{4.490} & \best{40.00} & \best{2.699} & \sbest{51.00} & \best{15.41} & \sbest{0.6049} & \best{0.6344}\\
        \addlinespace[2pt]\hdashline\addlinespace[2pt]
        $\Delta$ (Full$-$600) & +0.14 & +0.0024 & +0.0783 & $-$0.0142 & +2.00 & $-$0.225 & $-$2.45 & $-$0.051 & +1.80 & $-$2.25 & $-$0.0123 & +0.0124 \\
    \bottomrule
    \end{tabular}
    }
    \label{tab:overall_celeba}
\end{table}

\noindent\textbf{Full-data scaling (real-world benchmarks).}
Tab.~\ref{tab:overall_realworld} reports the LAFR 600-image vs.\ full FFHQ+FFHQR comparison on Wider-Test and WebPhoto-Test referenced in the main paper.

\begin{table}[H]
    \scriptsize
    \caption{LAFR trained on 600 FFHQ images vs.\ full FFHQ+FFHQR on Wider-Test and WebPhoto-Test. \best{Red}: best, \sbest{Blue}: second-best. $\Delta$ = Full $-$ 600 (signed); the changes are mixed in sign and modest in magnitude, indicating that scaling beyond 600 images yields limited and non-monotonic gains on real-world inputs.}
    \vspace{0.8mm}
    \centering
    \resizebox{\textwidth}{!}{%
    \begin{tabular}{l|*{5}{c}|*{5}{c}}
    \toprule
     & \multicolumn{5}{c|}{Wider-Test} & \multicolumn{5}{c}{WebPhoto-Test} \\
    \multirow{-2}{*}{\textbf{Training Set}} & CLIPIQA$\uparrow$ & MANIQA$\uparrow$ & MUSIQ$\uparrow$ & NIQE$\downarrow$ & FID$\downarrow$ & CLIPIQA$\uparrow$ & MANIQA$\uparrow$ & MUSIQ$\uparrow$ & NIQE$\downarrow$ & FID$\downarrow$ \\
    \midrule
    \textbf{Ours} (600) &
    \sbest{0.6330} & \sbest{0.6099} & \sbest{69.51} & \sbest{4.859} & \best{44.69} &
    \best{0.6956} & \sbest{0.6113} & \sbest{69.21} & \best{4.153} & \sbest{98.48} \\
    Full FFHQ+FFHQR &
    \best{0.6508} & \best{0.6336} & \best{72.13} & \best{4.135} & \sbest{62.15} &
    \sbest{0.6179} & \best{0.6251} & \best{70.65} & \sbest{4.768} & \best{91.73} \\
    \addlinespace[2pt]\hdashline\addlinespace[2pt]
    $\Delta$ (Full$-$600) &
    +0.0178 & +0.0237 & +2.62 & $-$0.724 & +17.46 &
    $-$0.0777 & +0.0138 & +1.44 & +0.615 & $-$6.75 \\
    \bottomrule
    \end{tabular}%
    }
    \label{tab:overall_realworld}
\end{table}

\section{Detailed Training-Set-Size Ablation}
\label{sec:supp_scale}

\begin{figure}[H]
    \centering
    \includegraphics[width=1\linewidth]{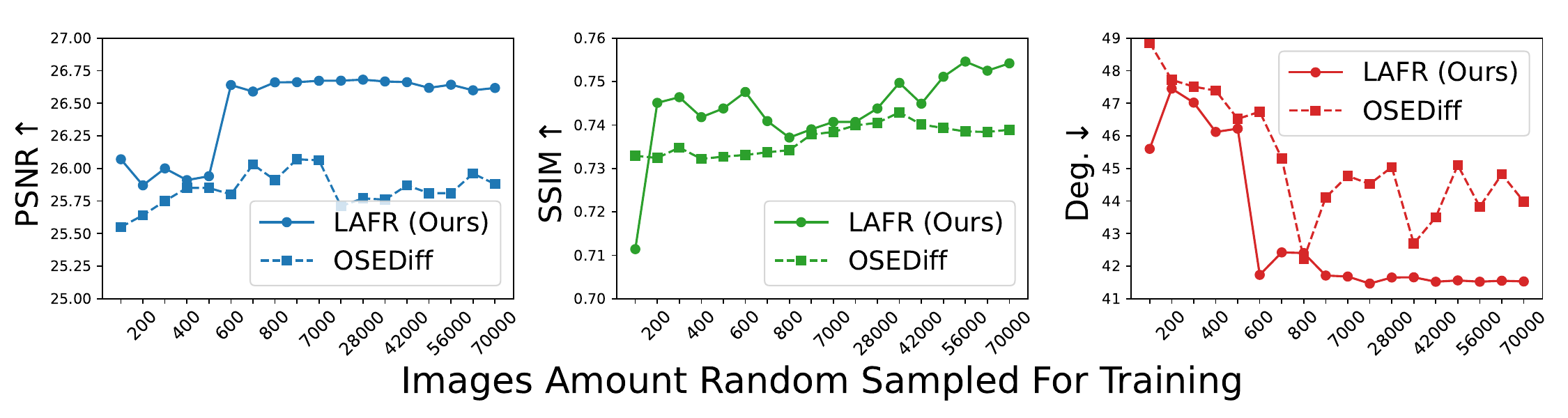}
    \caption{Performance trend of OSEDiff and our LAFR on CelebA-Test as the amount of training images increases. LAFR reaches competitive perceptual quality with only $\sim$600 images, whereas the OSEDiff baseline does not become competitive in the small-data regime, confirming that the proposed latent-codebook alignment makes face-domain adaptation data-efficient.}
    \label{fig:trend}
\end{figure}

The main paper summarises the data-efficiency trend in text (LAFR saturates around 600 images). Here we provide the full per-size measurements for both LAFR (Tab.~\ref{tab:ablation_scale}) and our re-trained OSEDiff baseline (Tab.~\ref{tab:ori_scale}) at training sizes ranging from 100 to 70,000 images, all randomly sampled from FFHQ, and visualise the trend in Fig.~\ref{fig:trend}. This sweep uses a single fixed configuration to isolate the effect of training-set size; its absolute numbers therefore differ slightly from the final headline model (which combines the alignment adapter, multilevel loss, and model simplification), and the 70K OSEDiff row equals the OSEDiff* entry of the main paper.

\begin{table}[H]
    \centering
    \caption{Detailed training-image-amount ablation for \textbf{our LAFR}, evaluated on CelebA-Test. Around $\sim$600 images, perceptual metrics (LPIPS, FID-related signals) saturate; scaling beyond 7K causes LPIPS regression due to the perception--distortion trade-off.}
    \resizebox{0.85\textwidth}{!}{%
    \begin{tabular}{c|cccccccc}
    \toprule
    \textbf{Training Images} & PSNR$\uparrow$ & SSIM$\uparrow$ & LPIPS$\downarrow$ & DISTS$\downarrow$ & MUSIQ$\uparrow$ & NIQE$\downarrow$ & Deg.$\downarrow$ & LMD$\downarrow$\\
    \midrule
    100   & 26.08 & 0.7384 & 0.2729 & 0.1918 & 71.17 & 5.117 & 46.11 & 3.023\\
    200   & 26.07 & 0.7114 & 0.2537 & 0.1765 & 71.94 & 4.817 & 45.60 & 2.883\\
    300   & 25.87 & 0.7451 & 0.2572 & 0.1836 & 71.28 & 4.881 & 47.45 & 2.948\\
    400   & 25.84 & 0.7425 & 0.2637 & 0.1841 & 71.02 & 4.872 & 46.75 & 2.932\\
    500   & 26.00 & 0.7464 & 0.2594 & 0.1910 & 71.37 & 5.080 & 47.02 & 2.824\\
    \textbf{600}   & 25.91 & 0.7418 & \textbf{0.2516} & 0.1782 & 71.33 & 4.754 & 46.12 & 2.810\\
    700   & 25.74 & 0.7388 & 0.2568 & 0.1786 & 72.20 & 4.841 & 46.22 & 3.047\\
    800   & 26.64 & 0.7476 & 0.2578 & 0.1794 & 68.64 & 4.901 & 41.73 & 2.819\\
    900   & 26.29 & 0.7309 & 0.2620 & 0.1810 & 69.39 & 4.805 & 42.42 & 2.795\\
    7,000 & 26.43 & 0.7461 & 0.2671 & 0.1831 & 69.37 & 5.044 & 42.87 & 2.773\\
    21,000 & 24.77 & 0.7328 & 0.3862 & 0.1937 & 70.99 & 4.855 & 53.44 & 3.600\\
    28,000 & 25.48 & 0.7497 & 0.3785 & 0.2032 & 66.82 & 5.342 & 51.02 & 3.194\\
    35,000 & 25.56 & 0.7449 & 0.3713 & 0.1962 & 70.64 & 5.183 & 50.56 & 3.176\\
    42,000 & 25.26 & 0.7390 & 0.3735 & 0.1912 & 70.27 & 5.254 & 50.52 & 3.172\\
    49,000 & 25.06 & 0.7371 & 0.3741 & 0.1910 & 71.25 & 4.968 & 52.05 & 3.347\\
    56,000 & 25.27 & 0.7407 & 0.3679 & 0.1826 & 71.50 & 4.953 & 49.65 & 3.144\\
    63,000 & 25.81 & 0.7438 & 0.3623 & 0.1819 & 69.84 & 4.852 & 46.71 & 2.901\\
    70,000 & 25.16 & 0.7380 & 0.3861 & 0.1973 & 70.06 & 5.148 & 52.04 & 3.368\\
    \bottomrule
    \end{tabular}
    }
    \label{tab:ablation_scale}
\end{table}

\begin{table}[H]
    \centering
    \caption{Detailed training-image-amount ablation for the \textbf{OSEDiff} baseline re-trained on FFHQ, evaluated on CelebA-Test. The 70K row is the OSEDiff* entry reported in the main paper. Unlike LAFR, OSEDiff does not reach competitive perceptual quality in the small-data regime and keeps relying on large-scale data.}
    \resizebox{0.85\textwidth}{!}{%
    \begin{tabular}{c|cccccccc}
    \toprule
    \textbf{Training Images} & PSNR$\uparrow$ & SSIM$\uparrow$ & LPIPS$\downarrow$ & DISTS$\downarrow$ & MUSIQ$\uparrow$ & NIQE$\downarrow$ & Deg.$\downarrow$ & LMD$\downarrow$\\
    \midrule
    100   & 25.95 & 0.7409 & 0.3719 & 0.1915 & 71.81 & 5.029 & 48.86 & 3.145\\
    200   & 26.04 & 0.7404 & 0.3607 & 0.1818 & 72.37 & 4.927 & 46.72 & 3.019\\
    300   & 26.15 & 0.7418 & 0.3592 & 0.1795 & 72.28 & 4.795 & 44.51 & 2.803\\
    400   & 26.25 & 0.7422 & 0.3524 & 0.1809 & 71.83 & 5.068 & 44.39 & 2.734\\
    500   & 26.25 & 0.7397 & 0.3545 & 0.1768 & 73.28 & 4.843 & 43.22 & 2.759\\
    600   & 26.15 & 0.7421 & 0.3560 & 0.1797 & 72.54 & 5.009 & 43.74 & 2.774\\
    700   & 26.20 & 0.7477 & 0.3460 & 0.1786 & 72.00 & 4.860 & 45.31 & 2.810\\
    800   & 26.43 & 0.7452 & 0.3472 & 0.1744 & 72.20 & 4.754 & 42.22 & 2.627\\
    900   & 26.31 & 0.7507 & 0.3478 & 0.1827 & 72.28 & 5.016 & 44.11 & 2.752\\
    7,000 & 26.47 & 0.7514 & 0.3394 & 0.1788 & 72.36 & 5.084 & 44.77 & 2.708\\
    21,000 & 26.11 & 0.7450 & 0.3480 & 0.1779 & 73.39 & 5.130 & 44.52 & 2.771\\
    28,000 & 26.17 & 0.7475 & 0.3387 & 0.1700 & 72.36 & 4.884 & 45.04 & 2.753\\
    35,000 & 26.56 & 0.7489 & 0.3377 & 0.1718 & 71.74 & 4.771 & 42.69 & 2.583\\
    42,000 & 26.57 & 0.7531 & 0.3397 & 0.1759 & 71.58 & 4.906 & 43.50 & 2.640\\
    49,000 & 26.21 & 0.7480 & 0.3511 & 0.1821 & 72.63 & 5.010 & 45.10 & 2.874\\
    56,000 & 26.21 & 0.7465 & 0.3402 & 0.1713 & 72.20 & 4.874 & 43.82 & 2.717\\
    63,000 & 26.36 & 0.7490 & 0.3428 & 0.1778 & 72.50 & 4.908 & 44.82 & 2.782\\
    \textbf{70,000} & 25.88 & 0.7388 & 0.3496 & 0.2278 & 69.98 & 5.328 & 67.40 & 5.408\\
    \bottomrule
    \end{tabular}
    }
    \label{tab:ori_scale}
\end{table}

\section{Why Not Use the Alignment Adapter Directly for Restoration?}
\label{sec:supp_why_not_direct}

Since the alignment adapter receives the latent code of an LQ input and outputs a corresponding ``HQ-aligned'' latent code, a natural question is: why not directly decode the aligned latent code as the final restoration, instead of feeding it through a diffusion model? We do not adopt this design for three reasons:

\begin{itemize}
    \item \textbf{Latent-level alignment $\neq$ pixel-level fidelity.} The adapter performs alignment only at the \emph{latent code} level, which does not guarantee image-level consistency or pixel-level restoration validity. The codebook is constrained to a $K$-entry vocabulary and a low-dimensional mapping, which is intentionally too restrictive to model the fine-grained texture, lighting, and identity-specific detail required for direct image output.
    \item \textbf{Limited representational capacity.} The codebook's alignment capability is bounded by its size and quantisation. It leverages the distributional similarity of facial images to learn an LQ$\to$HQ feature mapping efficiently, but does not encode the generative knowledge needed for de-novo facial detail synthesis.
    \item \textbf{Empirical evidence from a direct-VAE-training variant.} When we attempted to add restoration-related losses directly to the VAE and train it end-to-end, the VAE effectively degenerated into a UNet-like face restoration network. Prior literature has extensively shown that such architectures underperform diffusion-based restoration, supporting our choice to keep the VAE frozen and route generation through the diffusion UNet.
\end{itemize}

These observations motivate the two-stage design: the adapter corrects the \emph{starting point} of the diffusion trajectory, while the diffusion UNet still performs the actual high-quality face synthesis.

\section{Extended Reproducibility Checklist}
\label{sec:supp_repro}

This section expands the reproducibility block in the main paper into a complete checklist.

\noindent\textbf{Degradation pipeline.} We use the standard Real-ESRGAN-style second-order degradation, applied i.i.d.\ per training image:
\begin{itemize}
    \item Isotropic / anisotropic Gaussian blur with $\sigma\in[0.2,3.0]$, kernel size $\in\{7,9,\ldots,21\}$.
    \item Downsampling by a factor $r\in[1,8]$ via bilinear, bicubic, or area interpolation (uniform choice).
    \item Additive Gaussian noise with $\sigma_n\in[0,25]$ (uint8 scale) or Poisson noise (uniform 50/50 choice).
    \item JPEG compression with quality $q\in[30,95]$.
    \item Final resize to $512\times512$ via bicubic interpolation.
\end{itemize}
At evaluation on CelebA-Test we use a fixed $4\times$ degradation chain. Real-world benchmarks (WebPhoto, LFW, Wider) are used as provided.

\noindent\textbf{Training splits.} The 600-image FFHQ subset uses NumPy's \texttt{default\_rng(0)} with \texttt{choice} over the 70K indices without replacement; seeds $1$ and $2$ produce the additional subsets used for the robustness study (Tab.~\ref{tab:subset_robust}). The resulting integer ID lists, one file per seed ($\texttt{ffhq\_subset\_seed}k\texttt{.txt}$ for $k\in\{0,1,2\}$), are included in the supplementary archive.

\noindent\textbf{Evaluation set sizes.} CelebA-Test: 3\,000 images (standard split). WebPhoto-Test: 407 images. LFW-Test: 1\,711 images. Wider-Test: 970 images.

\noindent\textbf{Precision and hardware.} All training and inference use fp16 mixed precision on a single NVIDIA L40 (48\,GB). Stage-1 batch size 16, Stage-2 batch size 2.

\noindent\textbf{Timing protocol.} Inference time in the main paper's efficiency table (Tab.~3 there) is the full restoration pipeline, including VAE encode, adapter forward pass, one diffusion denoising step (for one-step methods) or the reported number of steps (multi-step methods), and VAE decode. Timings are the mean over 100 $512^{2}$ images with batch size 1 after a 10-step warm-up, measured with \texttt{torch.cuda.synchronize}.

\noindent\textbf{Metric conventions.} PSNR is reported on the Y channel after colour
correction of the restored image; skipping the colour correction or scoring on RGB shifts the
number by more than a decibel, so both settings must match when comparing against our
tables. LPIPS uses the AlexNet backbone unless a row is explicitly marked LPIPS-VGG, and FID
is computed with an InceptionV3 feature extractor against the reference split named in the
corresponding caption.

\noindent\textbf{Random seeds.} Global seed 0 unless otherwise noted; the subset-robustness study uses seeds 0, 1, 2 only for FFHQ subset sampling, with all other randomness fixed at 0.

\noindent\textbf{Hyperparameter summary.} Tab.~\ref{tab:supp_hparams} collects all hyperparameters of both stages in one place for reproduction.

\begin{table}[H]
    \centering
    \footnotesize
    \caption{Full hyperparameter summary for LAFR. Stage 1 trains the alignment adapter; Stage 2 fine-tunes the Conv-LoRA. ``--'' = not applicable.}
    \begin{tabular}{l|l|l}
    \toprule
    Hyperparameter & Stage 1 (adapter) & Stage 2 (LoRA) \\ \midrule
    Optimiser & AdamW & AdamW \\
    Learning rate & $1\mathrm{e}{-4}$ & $5\mathrm{e}{-5}$ \\
    LR schedule & cosine decay & constant \\
    Batch size & 16 & 2 \\
    Iterations / epochs & 100 epochs & 17K steps \\
    Weight decay & $1\mathrm{e}{-2}$ & $1\mathrm{e}{-2}$ \\
    Precision & fp16 & fp16 \\
    Trainable params & 5.2\,M (adapter) & 2.3\,M (Conv-LoRA) \\
    Codebook size $K$ / dim $d$ & $1024$ / $256$ & -- \\
    Soft-assign temperature $\tau$ & $0.07$ & -- \\
    LoRA rank & -- & 4 \\
    Loss & $\mathcal{L}_{\text{align}}$ ($\ell_1$ latent) & $\mathcal{L}_{\text{total}}$ (main Eq.~6) \\
    GPU & 1$\times$ L40 (48\,GB) & 1$\times$ L40 (48\,GB) \\
    \bottomrule
    \end{tabular}
    \label{tab:supp_hparams}
\end{table}

\noindent\textbf{Inference-cost breakdown.} The one-step inference time reported in the main paper (0.121\,s) decomposes into VAE encode, the adapter forward pass, a single UNet denoising step, and VAE decode. The adapter is negligible relative to the UNet step and VAE codec; the dominant cost is the single UNet evaluation, which is why LAFR is on par with the OSEDiff one-step baseline despite adding the adapter. Multi-step diffusion baselines pay this UNet cost once per step, explaining their order-of-magnitude slower inference in the main paper's efficiency table.

\section{Larger Qualitative Comparisons and Failure Cases}
\label{sec:supp_quals}

\begin{figure}[t]
    \centering
    \includegraphics[width=1.0\linewidth]{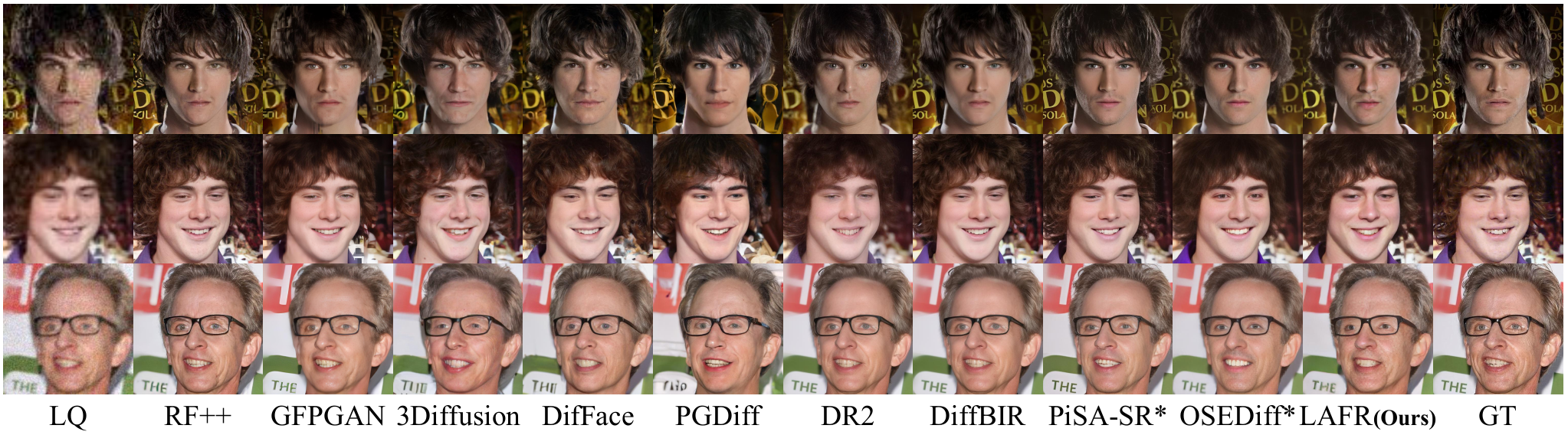}
    \caption{Full qualitative comparison on CelebA-Test against all baselines (the main paper shows a focused subset with crops), at full page width for detail-level inspection. RF++ = RestoreFormer++~\cite{wang2023restoreformerpp}.}
    \label{fig:supp_quals_celeba}
\end{figure}

\begin{figure}[t]
    \centering
    \includegraphics[width=1.0\linewidth]{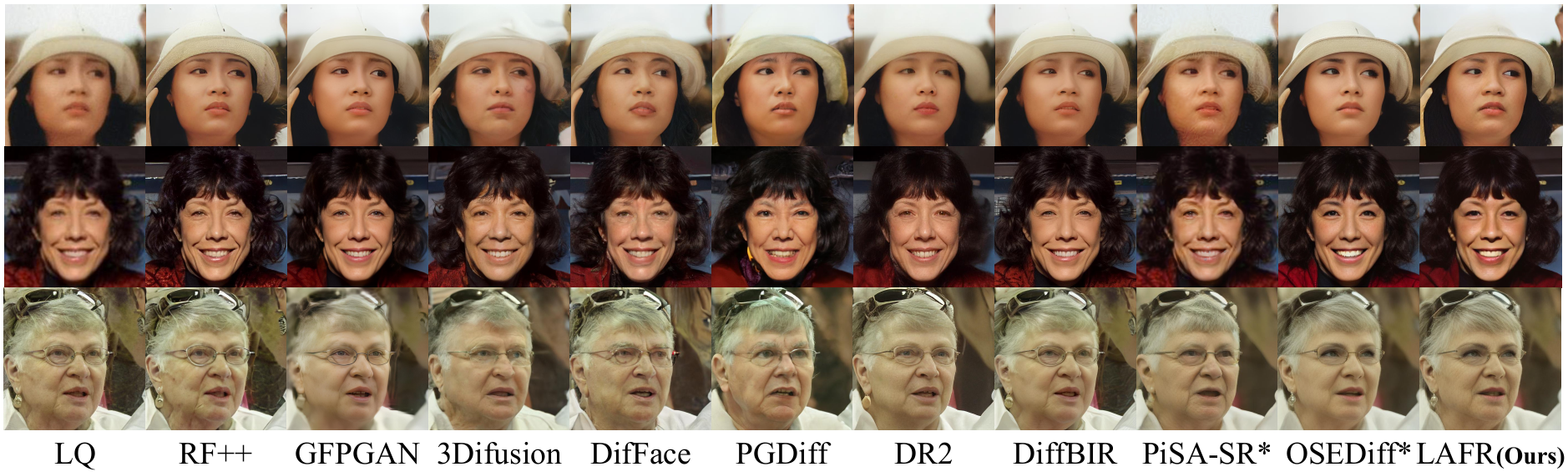}
    \caption{Large-scale qualitative comparisons on real-world benchmarks (WebPhoto-Test, LFW-Test, Wider-Test), at full page width for detail-level inspection.}
    \label{fig:supp_quals_real}
\end{figure}

\begin{figure}[t]
    \centering
    \includegraphics[width=0.72\linewidth]{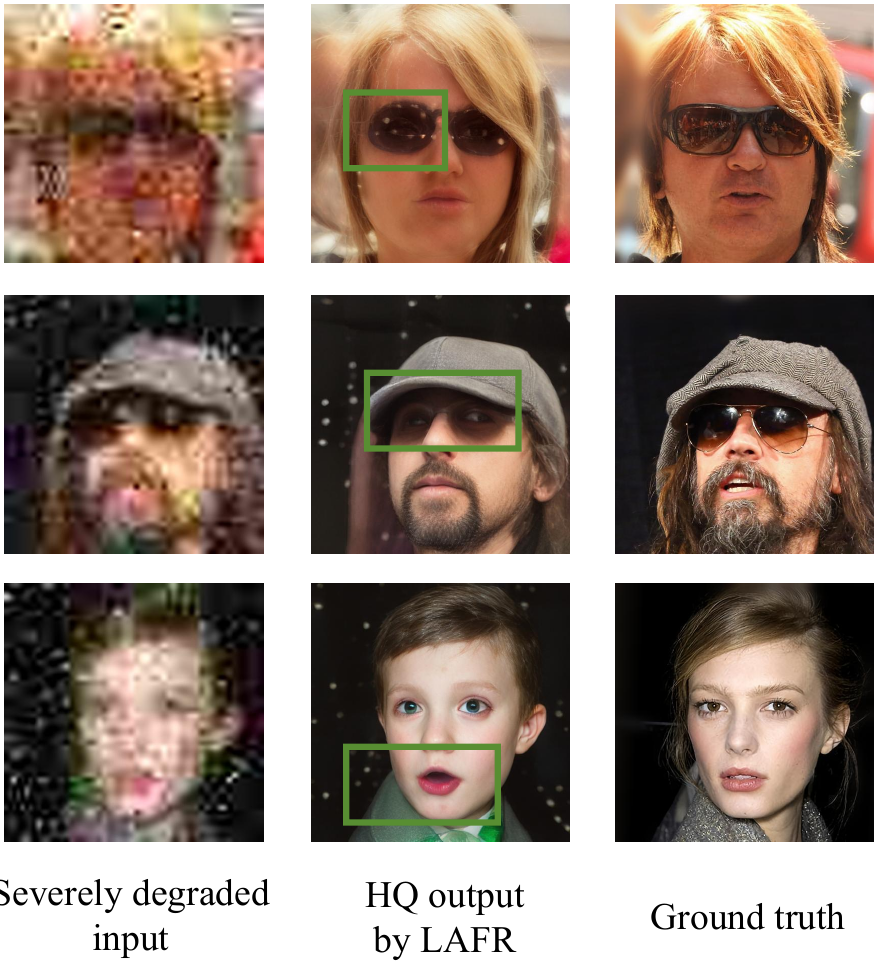}
    \caption{Failure cases on benchmark inputs. Columns: severely degraded input, LAFR output, and ground truth; green boxes mark the identity-sensitive regions (eyes, mouth) where LAFR deviates most from the GT. Under heavy occlusion (\eg sunglasses) and very severe degradation, LAFR synthesises plausible but GT-inconsistent eyes and mouth/teeth: such inputs leave the aligned-frontal, in-distribution regime, so the data-efficient model has limited generalisation budget.}
    \label{fig:supp_failure}
\end{figure}

The main paper (Fig.~5 there) shows a focused CelebA-Test comparison with fewer methods per row and zoomed crops of identity-sensitive regions. Here we complement it with the \emph{full} CelebA-Test comparison against \emph{all} baselines in Fig.~\ref{fig:supp_quals_celeba}, and the real-world comparison (WebPhoto-Test, LFW-Test, Wider-Test) in Fig.~\ref{fig:supp_quals_real}; both are shown at full page width so readers can zoom in for eye/mouth/skin detail. We also include a failure case in Fig.~\ref{fig:supp_failure}.

\noindent\textbf{Failure cases on benchmark inputs.} Fig.~\ref{fig:supp_failure} shows representative failures on severely degraded benchmark faces (no auxiliary dataset is used). Each row gives the severely degraded input, the LAFR output, and the ground truth; green boxes mark the identity-sensitive regions where the restoration deviates most from the GT. The failures concentrate on these regions under occlusion (\eg dark sunglasses) and very heavy degradation: the eyes behind opaque glasses cannot be recovered faithfully and the mouth/teeth acquire artifacts, because such inputs fall outside LAFR's aligned-frontal, in-distribution regime, where the codebook anchor and diffusion prior over-regularise toward a clean frontal face. This is consistent with the failure categories listed in the main paper (occlusion, large-yaw pose, and heavy JPEG${+}$blur).

\section{Extended Adapter Sensitivity Analysis}
\label{sec:supp_sensitivity}
Sec.~\ref{sec:supp_ablation_tables} ablates the core adapter design choices (full-latent vs.\ residual output, hard vs.\ soft inference, and codebook size $K$). Here we add the two remaining hyperparameters: the soft-assignment temperature $\tau$ and the descriptor dimension $d$ (Tab.~\ref{tab:supp_sensitivity}). Both exhibit a broad, flat optimum: FID is stable for $\tau\in[0.05,0.10]$ and degrades only for very small (over-peaked assignment, unstable gradients) or very large $\tau$ (over-smoothed mixture); $d{=}256$ is sufficient, with $d{=}512$ giving no meaningful gain at higher cost. We therefore use $\tau{=}0.07$ and $d{=}256$, and the method is not sensitive to small changes around these values.

\begin{table}[H]
    \centering
    \footnotesize
    \caption{Sensitivity to soft-assignment temperature $\tau$ and descriptor dimension $d$ on CelebA-Test (600-image training, all else fixed at the default). Default (ours): $\tau{=}0.07$, $d{=}256$. \best{Red}: best.}
    \begin{tabular}{l|cc @{\hskip 8pt} l|cc}
    \toprule
    $\tau$ & PSNR$\uparrow$ & FID$\downarrow$ & $d$ & PSNR$\uparrow$ & FID$\downarrow$ \\ \midrule
    0.03 & 26.10 & 18.40 & 128 & 26.12 & 18.85 \\
    \textbf{0.07} & \best{26.26} & \best{17.66} & \textbf{256} & \best{26.26} & \best{17.66} \\
    0.10 & 26.22 & 17.95 & 512 & 26.27 & 17.71 \\
    0.20 & 26.05 & 19.10 & -- & -- & -- \\
    \bottomrule
    \end{tabular}
    \label{tab:supp_sensitivity}
\end{table}

\section{Latent-Distribution Metrics: Methodology}
\label{sec:supp_latent_method}
This section details how the two latent-distribution distances reported in the main paper are computed, so the measurement is fully reproducible, and states the estimator used for the rank-deficient covariance.

\noindent\textbf{Latent extraction.} For each of $N{=}1000$ paired CelebA-Test images we encode, with the \emph{frozen} SD VAE, the HQ image, the LQ image, the LQ image through the LoRA-fine-tuned encoder, and the adapter-aligned latent, obtaining tensors in $\mathbb{R}^{4\times 64\times 64}$ that we flatten to $16384$-d vectors. All sets share the same image indices, so every comparison is paired. Every row of the main paper's table is a latent that \emph{enters denoising}; a method that leaves that latent untouched and instead conditions the UNet internally, as ControlNet-style injection does, has no entry in this space and is compared on restoration metrics instead.

\noindent\textbf{Why two statistics and not three.} We report exactly two: an RBF maximum mean discrepancy, which is non-parametric, and the squared Gaussian Wasserstein-2 distance $d_{\mu\Sigma}^{2}$, which is parametric. We do \emph{not} additionally report a ``latent FID''. For a single pair of fitted Gaussians the Fr\'echet distance
$\|\mu_X-\mu_Y\|_2^2+\operatorname{tr}\!\big(\Sigma_X+\Sigma_Y-2(\Sigma_X\Sigma_Y)^{1/2}\big)$
\emph{is} the squared Wasserstein-2 distance between them; the two names denote one quantity, so quoting both would be a restatement dressed as corroboration, and any numerical disagreement between them can only come from an unstated difference in scaling, projection or estimator.

\noindent\textbf{MMD.} We use an unbiased estimate of the squared maximum mean discrepancy with a Gaussian RBF kernel whose bandwidth $\sigma$ is set by the median heuristic (median pairwise distance over a pooled sample), averaged over kernels at $\{0.5,1,2\}\sigma$ for robustness. MMD requires no covariance estimate, which is precisely why it is reported next to the Gaussian statistic.

\noindent\textbf{Gaussian $d_{\mu\Sigma}^{2}$, and the rank deficiency it inherits.} $d_{\mu\Sigma}^{2}$ is the squared Wasserstein-2 distance between the fitted Gaussians. We report the \emph{total} only. It is decomposable in principle into a mean-shift term $\|\mu_X-\mu_Y\|_2^{2}$ and a covariance term, and it would be tempting to quote the two separately as evidence that the adapter corrects the shape of the distribution and not merely its location; we deliberately do not, because with the rank-deficient covariance estimate described below the split is not reliable enough to carry that argument. The estimator must be stated, because with $N{=}1000$ samples in $d{=}16384$ dimensions the empirical covariances have rank at most $N-1=999$ and are therefore singular by construction. The mean term is unaffected. For the covariance term we evaluate the trace on the subspace spanned by the pooled sample and add a small ridge $\epsilon I$ before the matrix square root for numerical stability, following the standard Fr\'echet-distance implementations; the product $(\Sigma_X\Sigma_Y)^{1/2}$ is obtained from the eigendecomposition of the symmetrised product restricted to its non-negligible eigenvalues. Two consequences should be kept in mind when reading the table. First, the absolute magnitude of $d_{\mu\Sigma}^{2}$ is estimator-dependent and is not comparable across papers; only the \emph{ordering} of rows, all computed under one fixed estimator on the same 1\,000 image indices, is meaningful. Second, a rank-deficient covariance estimate is biased, and the bias is shared across rows rather than cancelling, which is the second reason we report the non-parametric MMD beside it.


\subsection{Latent Interpolation Between the LQ and the Aligned Latent}
\label{supp:latent_interp}

The distances above characterise the two endpoints of the alignment. To show what the
adapter does \emph{between} them, and in particular to localise where alignment stops
helping, we walk the straight path
\begin{equation}
\label{eq:supp_interp}
z(\lambda)=(1-\lambda)\,z_{LQ}+\lambda\,z_{\text{aligned}},\qquad
\lambda\in\{0,0.25,0.5,0.75,1\},
\end{equation}
and decode every $z(\lambda)$ with the \emph{frozen} VAE decoder. Here $z_{LQ}$ is the
deterministic (mode) encoding of the LQ image by the frozen SD~v2-1 encoder and
$z_{\text{aligned}}$ is the Stage-1 adapter output. At each $\lambda$ we record the
$\ell_1$ distance and cosine similarity to the paired HQ latent $z_{HQ}$, a whitened
distance $d_{HQ}$ to the latent statistics of 200 held-out CelebA-HQ images, and LPIPS and
PSNR of the \emph{decoded} latent against the HQ image. Five cases are used: an ordinary
CelebA-Test face; the same face under much heavier degradation ($\sigma{=}8$, $r{=}8$,
$\delta{=}15$, $q{=}50$); two DIV2K crops (a crocodile and a church facade); and an
ImageNet animal crop. Tab.~\ref{tab:supp_interp} gives the trajectories.

\begin{table}[!htbp]
    \centering
    \footnotesize
    \caption{Latent interpolation $z(\lambda)=(1-\lambda)z_{LQ}+\lambda z_{\text{aligned}}$
    with the Stage-1 adapter output as the aligned endpoint. $\ell_1$ and $\cos$ are taken
    against the paired HQ latent $z_{HQ}$; $d_{HQ}$ is a whitened distance to the latent
    statistics of 200 held-out CelebA-HQ images; LPIPS and PSNR are computed on the
    \emph{decoded} latent against the HQ image and are diagnostics of where the latent sits,
    \emph{not} restoration scores --- decoding $z(\lambda)$ bypasses the diffusion UNet, so
    they are not comparable with the main paper's restoration results. $\lambda{=}0$ is the
    raw LQ latent and $\lambda{=}1$ the aligned latent.}
    \begin{tabular}{l|c|ccc|cc}
    \toprule
    Case & $\lambda$ & $\|z(\lambda){-}z_{HQ}\|_1\downarrow$ & $\cos(z(\lambda),z_{HQ})\uparrow$ & $d_{HQ}\downarrow$ & LPIPS$_{\text{dec}}\downarrow$ & PSNR$_{\text{dec}}\uparrow$ \\ \midrule
    \multirow{5}{*}{Face, ordinary}
      & 0    & 0.659 & 0.582 & 0.934 & 0.754 & 24.52 \\
      & 0.25 & 0.572 & 0.626 & 0.816 & 0.731 & 24.22 \\
      & 0.5  & 0.496 & 0.673 & 0.723 & 0.717 & 23.71 \\
      & 0.75 & 0.436 & 0.716 & 0.664 & 0.717 & 23.13 \\
      & 1    & \best{0.401} & \best{0.745} & \best{0.648} & 0.733 & 22.40 \\ \midrule
    \multirow{5}{*}{Face, heavy degr.}
      & 0    & 0.542 & 0.779 & 1.198 & 0.664 & 22.62 \\
      & 0.25 & 0.484 & 0.799 & 1.054 & 0.632 & 22.76 \\
      & 0.5  & 0.450 & 0.815 & 0.929 & 0.609 & 22.75 \\
      & 0.75 & \best{0.443} & \best{0.823} & 0.833 & \best{0.597} & 22.46 \\
      & 1    & 0.464 & 0.814 & \best{0.775} & 0.601 & 21.67 \\ \midrule
    \multirow{5}{*}{DIV2K 0895 (crocodile)}
      & 0    & 0.770 & 0.434 & 1.405 & 0.920 & 18.85 \\
      & 0.25 & 0.684 & 0.470 & 1.281 & 0.908 & 18.79 \\
      & 0.5  & 0.611 & 0.509 & 1.179 & 0.900 & 18.65 \\
      & 0.75 & 0.553 & 0.546 & 1.105 & \best{0.899} & 18.49 \\
      & 1    & \best{0.517} & \best{0.570} & \best{1.063} & 0.922 & 18.31 \\ \midrule
    \multirow{5}{*}{DIV2K 0879 (facade)}
      & 0    & 0.664 & 0.572 & 1.265 & 0.856 & 19.32 \\
      & 0.25 & 0.596 & 0.606 & 1.167 & 0.820 & 19.33 \\
      & 0.5  & 0.543 & 0.635 & 1.092 & 0.793 & 19.10 \\
      & 0.75 & 0.511 & \best{0.654} & 1.043 & \best{0.783} & 18.74 \\
      & 1    & \best{0.504} & 0.653 & \best{1.026} & 0.786 & 18.28 \\ \midrule
    \multirow{5}{*}{ImageNet animal}
      & 0    & 0.472 & 0.790 & 1.448 & 0.461 & 26.13 \\
      & 0.25 & 0.409 & 0.818 & 1.325 & 0.379 & 26.05 \\
      & 0.5  & 0.370 & 0.842 & 1.219 & \best{0.359} & 25.65 \\
      & 0.75 & \best{0.362} & \best{0.857} & 1.136 & 0.393 & 25.19 \\
      & 1    & 0.387 & 0.849 & \best{1.080} & 0.436 & 24.41 \\
    \bottomrule
    \end{tabular}
    \label{tab:supp_interp}
\end{table}

\IfFileExists{fig/latent_interp_decode_sml_new.pdf}{%
\begin{figure}[!htbp]
    \centering
    \includegraphics[width=\linewidth]{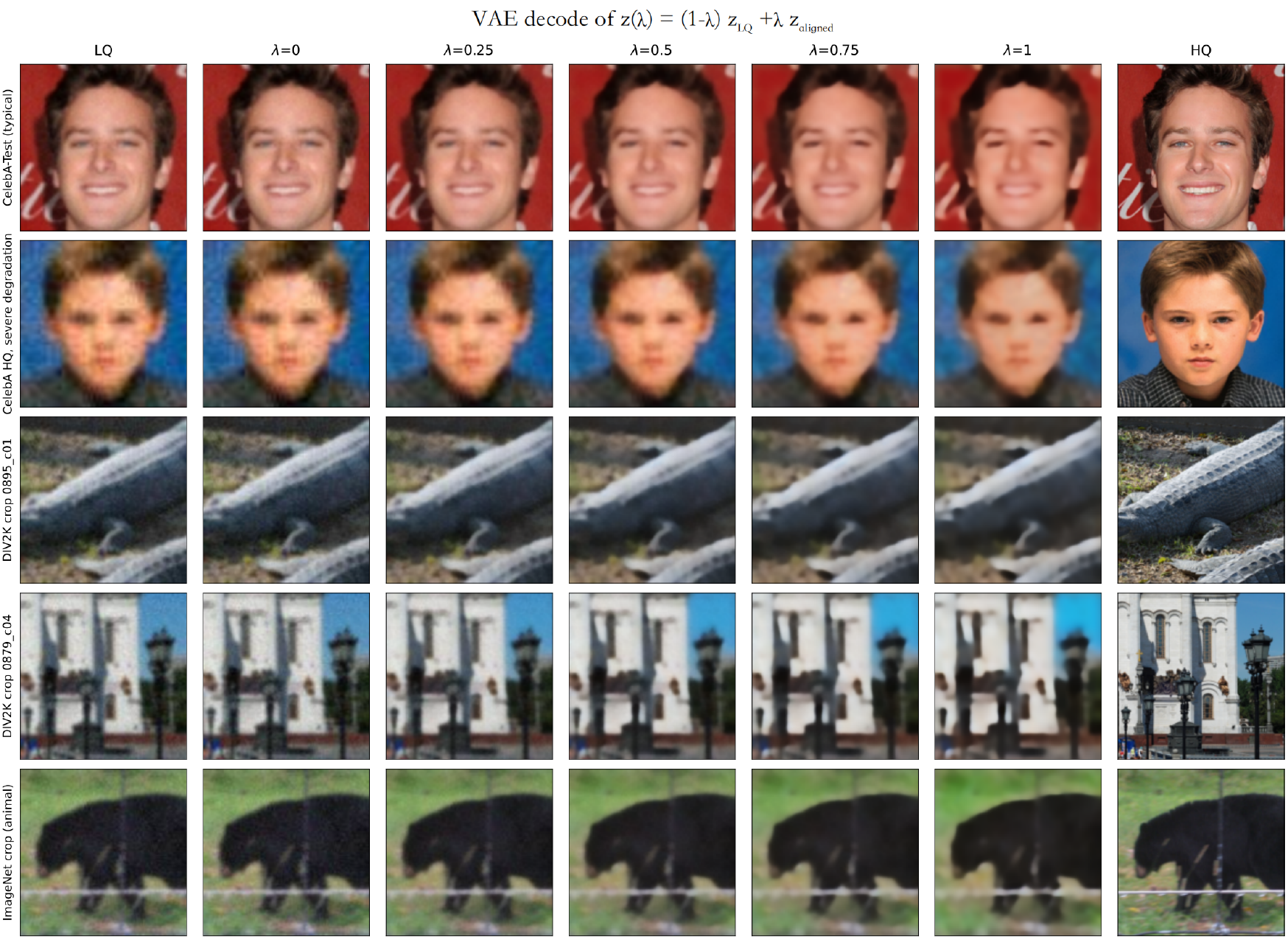}
    \caption{Decoded latent interpolation $z(\lambda)=(1-\lambda)z_{LQ}+\lambda
    z_{\text{aligned}}$ for the five cases of Tab.~\ref{tab:supp_interp}, decoded with the
    frozen VAE decoder; the LQ input and the HQ target are shown on either side for
    reference. On the in-domain faces, noise and compression artifacts are shed
    progressively along the walk; on the natural-image cases the walk moves the latent
    toward the face statistics without making the decoded image better.}
    \label{fig:supp_interp_decode}
\end{figure}}{}

\noindent\textbf{In-domain faces align monotonically.} For the ordinary face the $\ell_1$
distance to $z_{HQ}$ falls from $0.659$ to $0.401$ ($-39\%$), the cosine rises from $0.582$
to $0.745$, and $d_{HQ}$ drops from $0.934$ to $0.648$. The decoded images shed noise and
compression artifacts progressively and the decoded LPIPS improves from $0.754$ to $0.717$,
with its optimum at $\lambda\in[0.5,0.75]$. Decoded PSNR falls over the same walk
($24.52\to22.40$), which is expected rather than a regression: alignment removes the
degradation-specific high frequencies that a per-pixel metric still rewards.

\noindent\textbf{Where alignment breaks down.} On the heavily degraded face the trajectory
is no longer monotone. The $\ell_1$ distance bottoms out at $\lambda{=}0.75$ ($0.443$) and
rises again at $\lambda{=}1$ ($0.464$), and the decoded LPIPS has the same shape
($0.597\to0.601$). The adapter therefore \emph{overshoots on the final segment} once the
input leaves the degradation range it was fitted on. This is a concrete localisation of the
failure mode that the main paper reports qualitatively, and it is consistent with the
codebook statistics below, where this case sits farthest from the training distribution.

\noindent\textbf{Out-of-domain inputs are pulled toward the face manifold without
improving.} For \emph{every} natural-image case $d_{HQ}$ decreases monotonically (the
adapter maps any input toward the face latent statistics, as it must, since its cells cover
only face appearance modes), yet the decoded images do not get better: PSNR falls
monotonically (crocodile $18.85\to18.31$, facade $19.32\to18.28$) and LPIPS worsens again at
$\lambda{=}1$. This is the mechanism behind the domain limitation of
Sec.~\ref{sec:supp_natural}, observed directly along the latent path rather than inferred
from the outputs.

\noindent\textbf{Codebook distance measures distance from the training distribution, not
from ``being a face''.} The mean quantisation distance is smallest for the ordinary face
($2.60$) and larger for the natural crops ($3.47$--$3.98$), which also activate more entries
($293$--$353$ against $245$); but the heavily degraded \emph{face} is farther still ($6.28$).
Tab.~\ref{tab:supp_interp_codebook} reports the per-case statistics, and we state the
interpretation this carefully for that reason.

\noindent\textbf{Why the correction has to happen at the entrance.} Repeating the sweep on
the conditioning latent that the deployed one-step model actually consumes
(Tab.~\ref{tab:supp_interp_onestep}) yields a usable restoration only at $\lambda{=}1$
(LPIPS $0.373$, PSNR $23.5$\,dB on the ordinary face); it collapses as soon as the
conditioning is moved toward the unaligned end ($\lambda{=}0$: LPIPS $0.797$, PSNR
$13.8$\,dB). A one-step model has no later sampling steps in which to repair a misaligned
conditioning latent, which is precisely why the correction belongs \emph{before} denoising
rather than inside it; the sensitivity is a property of the one-step regime, not of the
adapter.

\begin{table}[!htbp]
    \centering
    \footnotesize
    \caption{Sensitivity of the deployed \emph{one-step} restoration to its conditioning
    latent. Each $z(\lambda)$ of Eq.~\eqref{eq:supp_interp} is fed to the one-step model in
    place of its usual conditioning latent, whose value is the $\lambda{=}1$ endpoint. Cells
    are LPIPS$\downarrow$\,/\,PSNR$\uparrow$ (dB) against the HQ image. These are
    single-image values on the five cases, not benchmark averages. A usable output exists
    only at the aligned endpoint: a one-step model has no later sampling steps in which to
    repair a misaligned conditioning latent.}
    \begin{tabular}{l|ccccc}
    \toprule
    Case & $\lambda{=}0$ & $\lambda{=}0.25$ & $\lambda{=}0.5$ & $\lambda{=}0.75$ & $\lambda{=}1$ \\ \midrule
    Face, ordinary          & 0.797 / 13.8 & 0.806 / 13.6 & 0.809 / 13.6 & 0.758 / 14.8 & \best{0.373 / 23.5} \\
    Face, heavy degr.       & 0.604 / 12.6 & 0.614 / 12.7 & 0.631 / 13.2 & 0.593 / 14.7 & \best{0.259 / 22.3} \\
    DIV2K 0895 (crocodile)  & 0.826 / 11.9 & 0.830 / 11.9 & 0.841 / 11.9 & 0.871 / 12.8 & \best{0.728 / 18.7} \\
    DIV2K 0879 (facade)     & 0.764 / 12.1 & 0.772 / 12.1 & 0.780 / 12.3 & 0.796 / 13.2 & \best{0.398 / 18.6} \\
    ImageNet animal         & 0.572 / 11.6 & 0.562 / 11.9 & 0.549 / 13.0 & 0.521 / 15.5 & \best{0.346 / 24.2} \\
    \bottomrule
    \end{tabular}
    \label{tab:supp_interp_onestep}
\end{table}

\begin{table}[!htbp]
    \centering
    \footnotesize
    \caption{Codebook behaviour per case ($K{=}1024$). ``Quant.\ dist.'' is the mean
    distance from the descriptor to its selected entry, ``Active'' the number of distinct
    entries selected, and ``Entropy'' the entropy of the selection histogram. The quantity
    tracks distance from the \emph{training distribution}, not distance from ``being a
    face'': the heavily degraded face is farther out than any natural-image crop.}
    \begin{tabular}{l|ccc}
    \toprule
    Case & Quant.\ dist.$\downarrow$ & Active codes & Entropy \\ \midrule
    Face, ordinary          & \best{2.60} & 245 & 4.45 \\
    Face, heavy degr.       & 6.28 & 335 & 4.90 \\
    DIV2K 0895 (crocodile)  & 3.47 & 336 & 4.89 \\
    DIV2K 0879 (facade)     & 3.98 & 353 & 4.93 \\
    ImageNet animal         & 3.59 & 293 & 4.50 \\
    \bottomrule
    \end{tabular}
    \label{tab:supp_interp_codebook}
\end{table}

\IfFileExists{fig/latent_interp_onestep_new.pdf}{%
\begin{figure}[!htbp]
    \centering
    \includegraphics[width=\linewidth]{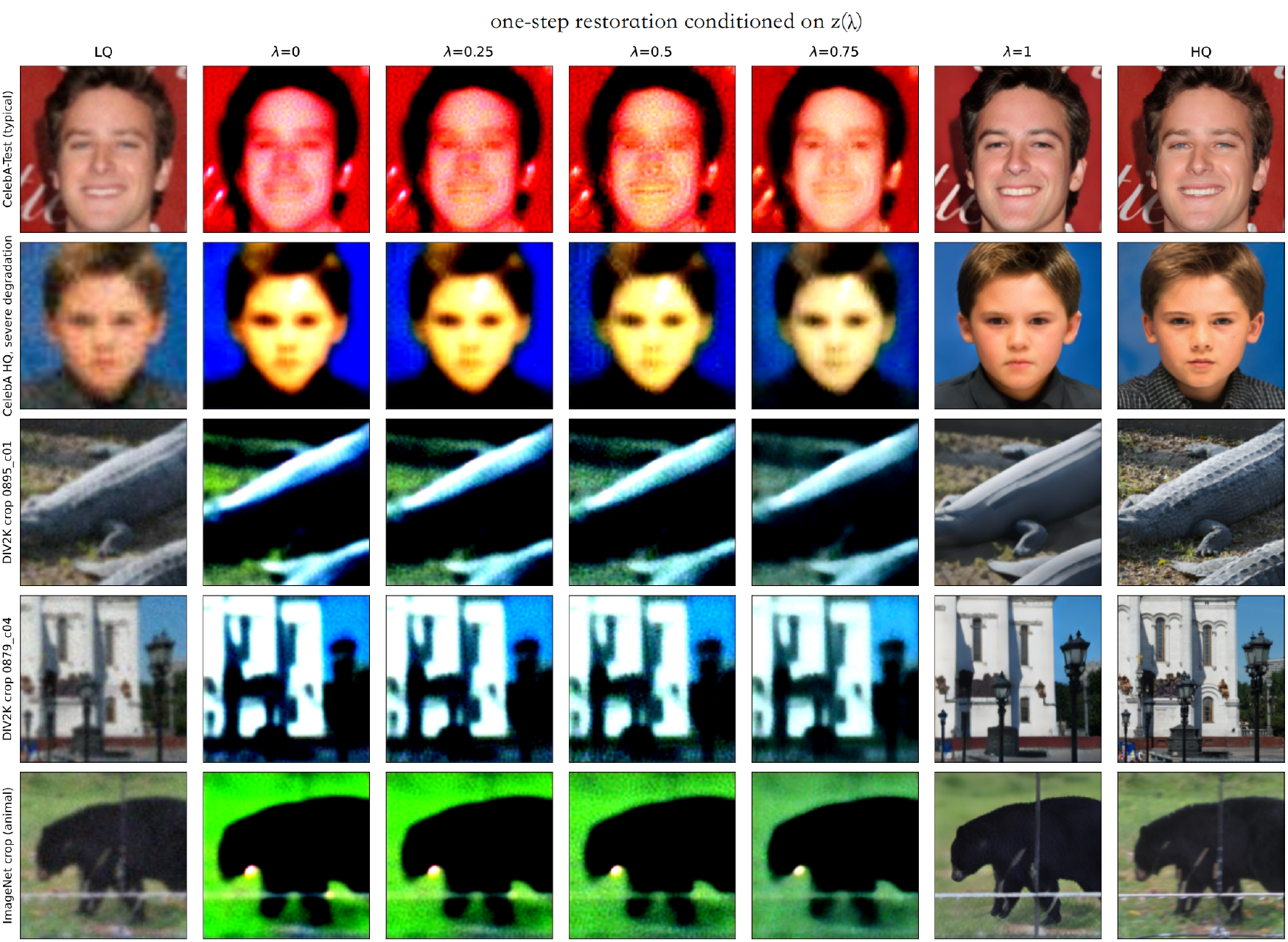}
    \caption{One-step restorations obtained by feeding $z(\lambda)$ to the deployed one-step
    model as its conditioning latent (numbers in Tab.~\ref{tab:supp_interp_onestep}); the LQ
    input and HQ target flank the sweep. The output is usable only at the aligned endpoint
    $\lambda{=}1$: for every $\lambda<1$ the single denoising step diverges into a saturated
    colour cast, which is what the ${\approx}14$\,dB PSNR values in
    Tab.~\ref{tab:supp_interp_onestep} record. A one-step model has no later step in which
    to recover from a misaligned conditioning latent.}
    \label{fig:supp_interp_onestep}
\end{figure}}{}

\IfFileExists{fig/latent_interp_curves_sml.pdf}{%
\begin{figure}[!htbp]
    \centering
    \includegraphics[width=\linewidth]{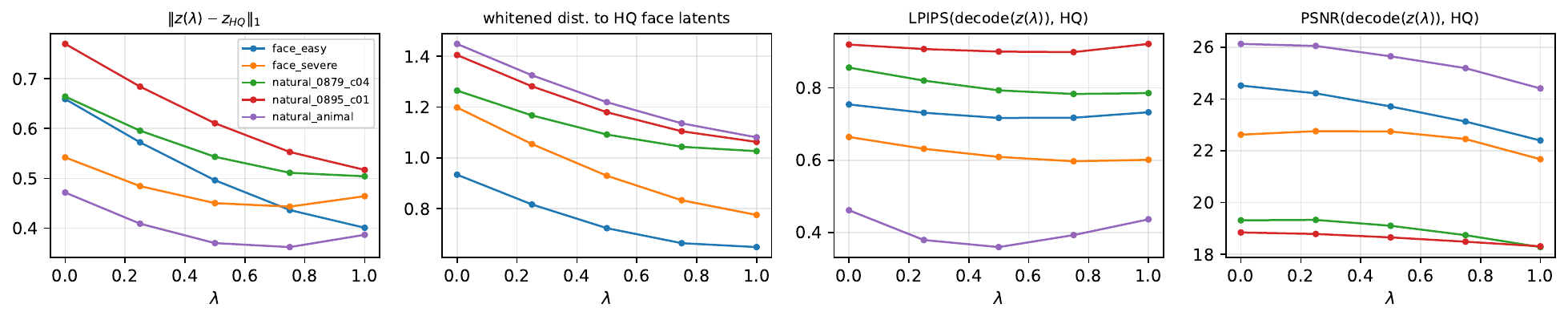}
    \caption{Interpolation trajectories of Tab.~\ref{tab:supp_interp} plotted against
    $\lambda$: $\ell_1$ distance to $z_{HQ}$, whitened distance $d_{HQ}$ to the HQ face
    statistics, and LPIPS and PSNR of the decoded latent. The whitened distance falls
    monotonically for every case, including the natural-image ones, whereas the decoded
    LPIPS turns back up before $\lambda{=}1$ on the heavily degraded face, the overshoot
    discussed in the text.}
    \label{fig:supp_interp_curves}
\end{figure}}{}

\noindent\textbf{Reading the table.} In the main paper's latent-distance table, the HQ-vs-HQ split provides a non-zero lower bound (finite-sample noise). The raw LQ latents are more than an order of magnitude farther from HQ than this bound on both measures, and the aligned latents reduce both by more than $4\times$, landing close to (though not at) the lower bound, consistent with the t-SNE visualisation and with the restoration gains.

\section{Identity Evaluation: Protocol and Discussion}
\label{sec:supp_identity}
The main paper reports identity evidence beyond the ArcFace degradation score (Deg.): ArcFace cosine similarity to GT, face-verification accuracy, and identity consistency to the LQ input. We give the exact protocol here.

\noindent\textbf{Embedding and similarity.} We use a fixed, publicly available ArcFace~\cite{deng2019arcface} recogniser (never fine-tuned, and \emph{never} used as a training loss; see the metric-leakage analysis in the main paper). Faces are aligned and cropped with the standard 5-point pipeline before embedding; similarity is the cosine between $\ell_2$-normalised embeddings.

\noindent\textbf{Verification accuracy (CelebA-Test).} For each restored/GT pair we accept a match when the cosine similarity exceeds the conventional threshold $0.6$; accuracy is the fraction of correctly accepted genuine pairs. This is a stricter, decision-level summary than the mean similarity and is reported alongside it.

\noindent\textbf{Identity consistency to LQ (real-world).} On WebPhoto and Wider, where no GT exists, we report the ArcFace cosine between the restored output and the (degraded) LQ input. Higher is better here in a \emph{relative} sense: a restoration that drifts to an unrelated identity, even if it looks clean, scores low, so this metric penalises identity hallucination that no-reference image-quality metrics would miss. It should be read jointly with the quality metrics, not as an absolute identity guarantee (the LQ input itself is degraded).

\noindent\textbf{On the absence of a human study.} We report no human perceptual study, and we do not treat CLIPIQA, MANIQA, MUSIQ or NIQE as substitutes for one. On the real-world benchmarks these no-reference scores and FID are read jointly as complementary indicators of competitiveness, never as evidence of preference, and the main paper's wording is chosen accordingly. The identity measurements above are the part of the real-world evaluation that is not opinion-based: ID-consistency penalises a restoration that drifts to a different person even when it looks clean, which is precisely the failure a no-reference quality score cannot see. A controlled two-alternative forced-choice study against the strongest baselines remains the right way to settle the perceptual comparison and is left for future work.

\noindent\textbf{Why not also train with ArcFace.} As shown in the main paper's loss ablation, optimising an ArcFace-based identity loss minimises the ArcFace-based Deg.\ metric (metric leakage) while harming perceptual and structural quality, because ArcFace features are trained to be invariant to lighting, pose, and texture, which are the very cues restoration must recover. We therefore keep ArcFace strictly for evaluation.

\section{Behaviour Outside the Face Domain}
\label{sec:supp_natural}

A recurring question is whether LAFR transfers to natural images. It does not, by
construction, and we document the failure mode explicitly rather than leaving it
implicit in the ``face restoration'' framing.

\begin{figure}[!htbp]
    \centering
    \includegraphics[width=0.62\linewidth]{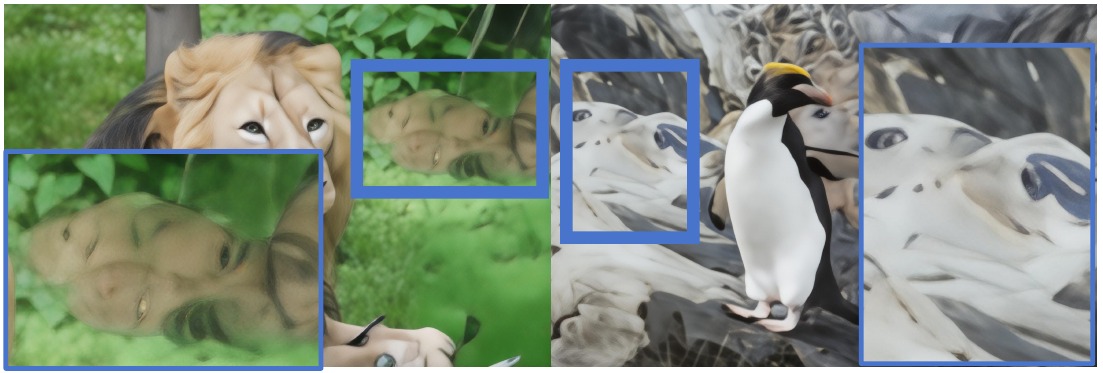}
    \caption{LAFR applied unmodified to non-face content. Because the alignment adapter can
    only return a face appearance mode, the diffusion pass renders ghostly human-face
    structure into the surrounding foliage and rock texture (blue boxes). This is the
    domain limitation stated in the main paper: LAFR is a face-domain method, and its
    data efficiency rests on exactly the regularity that natural images lack.}
    \label{fig:supp_natural}
\end{figure}

The codebook is fitted to 600 aligned faces, so its entries span face appearance modes
only. Applied unmodified to natural-image crops, the adapter maps every input onto the
nearest \emph{face} mode, and the diffusion pass then renders that anchor faithfully:
recognisable human-face structure (eyes, brows, a nose bridge) emerges in regions
of foliage and rock texture that contained none. Fig.~\ref{fig:supp_natural} shows
representative examples.

This is the expected consequence of the compactness measurement in the main paper
(Fig.~2): ResNet-50 features of ImageNet images have intra-class distance $24.13$ against
$13.28$ for CelebA, so a 1024-entry codebook estimated from 600 faces cannot cover the
natural-image manifold. Sec.~\ref{sec:supp_proof} makes the requirement quantitative: the
number of cells must grow with the number of appearance modes, at a rate set by the
intrinsic dimension of the support. For open-domain restoration that mode count is orders
of magnitude larger, so the very property that makes LAFR data-efficient on faces, namely strong geometric and
photometric regularity, is absent, and the method should not be
expected to transfer. We therefore treat open-domain restoration as out of scope rather
than as a deficiency to be closed by enlarging $K$. The remainder of this section makes that
statement quantitative.

\subsection{Quantitative Measurement on Natural Images}
\label{supp:div2k}

We take the 100 DIV2K validation images and extract $1\,000$ crops of $512\times512$ (ten
random positions per image, fixed seed), then synthesise LQ inputs with LAFR's \emph{own}
blind-face degradation operator, replacing its random ranges by three fixed levels so that
the measurement is reproducible:
\begin{equation}
y=\mathrm{JPEG}_q\!\big((x\otimes k_\sigma)\!\downarrow_r+\,n_\delta\big),
\end{equation}
followed by a bilinear resize back to $512^2$, with mild
$(\sigma,r,\delta,q)=(2,2,5,90)$, medium $(5,4,10,70)$ and severe $(8,8,15,50)$. Both models
receive byte-identical LQ inputs and the same cached prompt, are run at $\times 1$ scale with
the same wavelet colour correction, and FID is computed over the $1\,000$ restorations
against the $1\,000$ ground-truth crops. The LAFR rows use the same configuration as
Sec.~\ref{supp:latent_interp}, that is, our Stage-2 model together with the Stage-1 adapter described there. Tab.~\ref{tab:supp_div2k} reports the result, with
the unrestored LQ input included as a reference point.

\begin{table}[!htbp]
    \centering
    \footnotesize
    \caption{LAFR applied \emph{unmodified} to natural images: $1\,000$ DIV2K crops of
    $512^2$ at three fixed degradation levels. ``OSEDiff (general)'' is the publicly released
    general-purpose one-step model --- \emph{not} the FFHQ-retrained OSEDiff* of the main
    paper --- and ``OSEDiff-face'' is the released face-specialised model. All rows use
    byte-identical LQ inputs, the same prompt, and the same wavelet colour correction with
    PSNR-Y measured on the Y channel, matching the main paper's protocol. FID is computed
    against the $1\,000$ ground-truth crops. \best{Red}: best per block; the LQ row is a
    reference point and is excluded from ranking.}
    \resizebox{\textwidth}{!}{%
    \begin{tabular}{l|l|cccccccccc}
    \toprule
    Level & Method & FID$\downarrow$ & LPIPS$\downarrow$ & LPIPS-V$\downarrow$ & DISTS$\downarrow$ & PSNR$\uparrow$ & PSNR-Y$\uparrow$ & SSIM$\uparrow$ & NIQE$\downarrow$ & C-IQA$\uparrow$ & M-IQ$\uparrow$ \\ \midrule
    \multirow{3}{*}{mild}
      & \pnum{LQ input (no restoration)} & \pnum{56.95} & \pnum{0.4839} & \pnum{0.4267} & \pnum{0.2648} & \pnum{25.52} & \pnum{27.20} & \pnum{0.6805} & \pnum{6.43} & \pnum{0.3172} & \pnum{25.11} \\
      & LAFR, unmodified                 & 67.14 & 0.3585 & 0.4263 & 0.2307 & 23.12 & 24.78 & 0.6124 & 5.68 & 0.5711 & 59.00 \\
      & OSEDiff (general)                & \best{42.21} & \best{0.2469} & \best{0.3406} & \best{0.1689} & \best{23.26} & \best{24.90} & \best{0.6345} & \best{4.88} & \best{0.7136} & \best{68.77} \\ \midrule
    \multirow{4}{*}{medium}
      & \pnum{LQ input (no restoration)} & \pnum{148.48} & \pnum{0.7731} & \pnum{0.6176} & \pnum{0.3890} & \pnum{22.12} & \pnum{23.83} & \pnum{0.5397} & \pnum{10.59} & \pnum{0.3012} & \pnum{14.97} \\
      & LAFR, unmodified                 & 86.29 & 0.4358 & 0.4811 & 0.2688 & 22.31 & 23.99 & 0.5527 & \best{5.68} & 0.5011 & 53.53 \\
      & OSEDiff (general)                & \best{59.29} & \best{0.3403} & \best{0.4246} & \best{0.2209} & 22.15 & 23.80 & 0.5364 & 4.97 & \best{0.6624} & \best{65.53} \\
      & OSEDiff-face (released)          & 84.58 & 0.4199 & 0.4672 & 0.2728 & \best{22.37} & \best{24.04} & \best{0.5541} & 5.91 & 0.5427 & 56.37 \\ \midrule
    \multirow{3}{*}{severe}
      & \pnum{LQ input (no restoration)} & \pnum{221.10} & \pnum{0.8675} & \pnum{0.7222} & \pnum{0.4587} & \pnum{20.38} & \pnum{22.17} & \pnum{0.5043} & \pnum{14.53} & \pnum{0.2031} & \pnum{15.61} \\
      & LAFR, unmodified                 & 124.35 & \best{0.5178} & \best{0.5440} & \best{0.3161} & \best{20.75} & \best{22.53} & \best{0.5069} & 5.73 & 0.4934 & 51.69 \\
      & OSEDiff (general)                & \best{119.14} & 0.5440 & 0.5538 & 0.3333 & 20.08 & 21.82 & 0.4355 & \best{4.71} & \best{0.5907} & \best{55.11} \\
    \bottomrule
    \end{tabular}%
    }
    \label{tab:supp_div2k}
\end{table}

\noindent\textbf{At mild and medium degradation the ordering seen on face benchmarks
reverses.} The general-purpose one-step baseline leads on every metric: FID $42.21$ against
$67.14$ and LPIPS $0.247$ against $0.359$ at the mild level, FID $59.29$ against $86.29$ and
LPIPS $0.340$ against $0.436$ at the medium level. Once the input contains no face, a
face-specific prior is a liability rather than an advantage, which is what
Sec.~\ref{sec:supp_proof} predicts for a quantiser whose cells cover only face appearance
modes, since every input is routed to the nearest face mode whether or not that mode is
appropriate.

\noindent\textbf{At severe degradation the gap does not widen; it changes character.} We
flag this explicitly because it is the opposite of what one might expect. FID becomes
comparable ($124.35$ against $119.14$) and LAFR is in fact \emph{better} on every
fidelity-oriented metric (LPIPS $0.518$ vs.\ $0.544$, DISTS $0.316$ vs.\ $0.333$, PSNR-Y
$22.53$ vs.\ $21.82$, SSIM $0.507$ vs.\ $0.436$), trailing only on the no-reference scores
(CLIPIQA $0.493$ vs.\ $0.591$, MUSIQ $51.69$ vs.\ $55.11$). The two priors fail in different
ways: the face prior degrades gracefully by over-smoothing, whereas the general model keeps
synthesising sharp but incorrect texture. Read on their own, the no-reference metrics would
report the reverse ranking, a concrete instance of the caution stated in the main paper
that such scores are complementary indicators and not preference measures.

\noindent\textbf{This is a property of face-specialised one-step models, not of our
adapter.} The publicly released OSEDiff-face model behaves the same way at the medium level
(FID $84.58$, LPIPS $0.420$, DISTS $0.273$), tracking LAFR rather than the general baseline.
The limitation therefore follows from specialising a one-step prior to the face domain
(the same specialisation that buys the data efficiency reported in the main paper), and not
from the codebook adapter in particular. This is the sense in which we regard open-domain
restoration as outside the scope of the method rather than as a defect of it.

\IfFileExists{fig/div2k_qualitative.pdf}{%
\begin{figure}[!htbp]
    \centering
    \includegraphics[width=\linewidth]{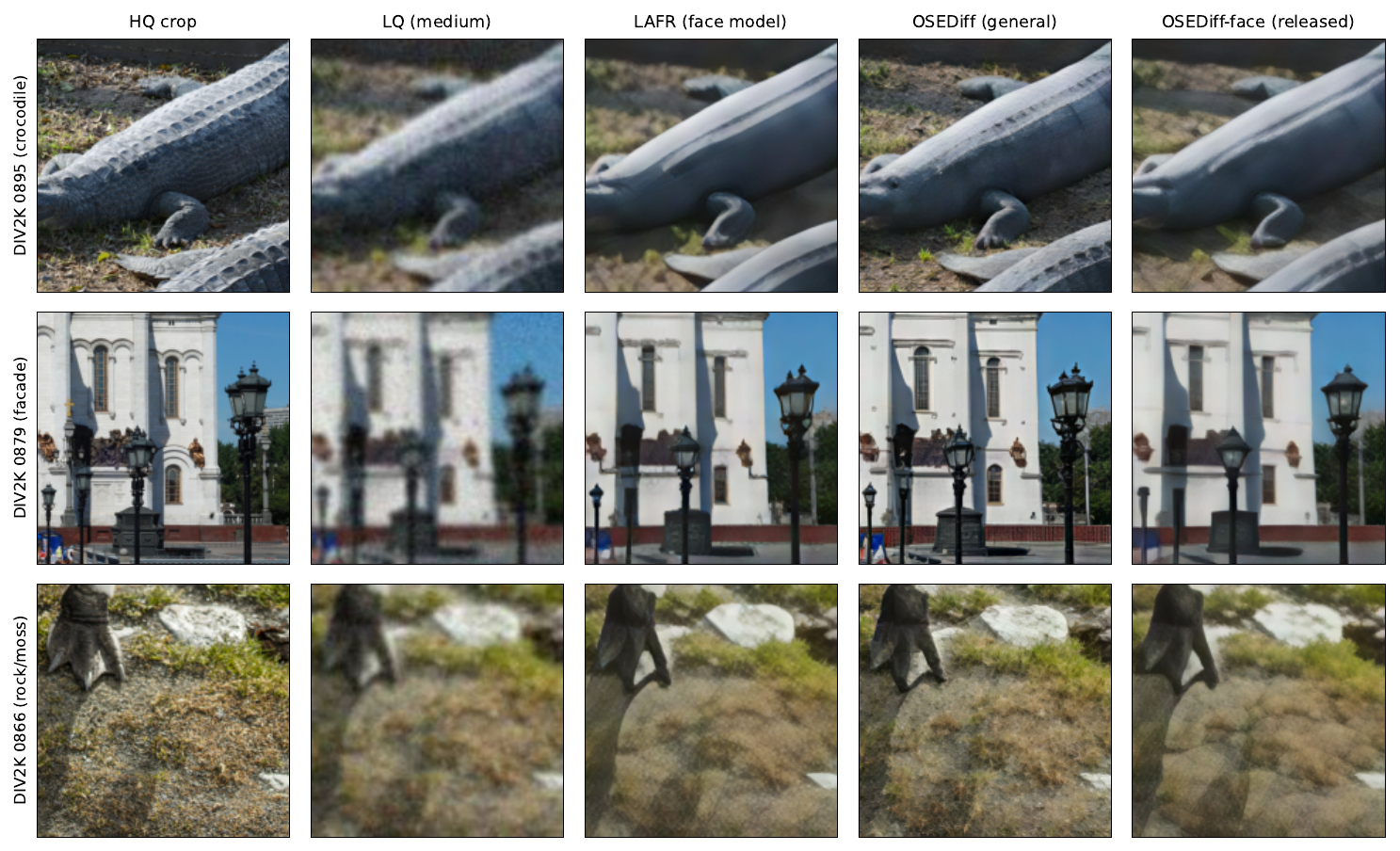}
    \caption{Qualitative comparison on DIV2K crops at the \emph{medium} degradation level.
    Both face-specialised models (ours and the released OSEDiff-face) flatten
    crocodile scales, brickwork and rock/moss rather than reconstructing them, while the
    general-purpose one-step model recovers texture. The visual similarity of the two
    face-specialised columns is the qualitative counterpart of their near-identical metrics
    in Tab.~\ref{tab:supp_div2k}, and supports reading this as a property of face
    specialisation rather than of our adapter.}
    \label{fig:supp_div2k_qual}
\end{figure}}{}

\section{Anticipated Questions and Additional Discussion}
\label{sec:supp_qa}
We collect here the questions most likely to arise, with concise answers grounded in the paper's experiments.

\noindent\textbf{Does the adapter generalise across SD versions / VAEs?} The adapter operates purely on the VAE latent ($4{\times}64{\times}64$ for SD~v2-1) and treats the VAE and UNet as frozen black boxes. Porting to another latent diffusion backbone only requires matching the adapter input/output to that backbone's latent shape and re-running the lightweight Stage-1 alignment; nothing in the design is specific to SD~v2-1 beyond the latent dimensions.

\noindent\textbf{Why not jointly train the adapter and the UNet?} Decoupling keeps the pretrained generative prior intact and makes the alignment stage independently trainable and inspectable (e.g., the latent-distance diagnostics of Sec.~\ref{sec:supp_latent_method}). Our head-to-head comparison against FaithDiff's jointly-retrained alignment (Tab.~\ref{tab:val_outside}) shows the decoupled adapter matches or beats it at a fraction of the parameters and inference cost, so joint training is not necessary for quality and is costlier and less stable.

\noindent\textbf{Is the 600-image result just overfitting?} No. The full-FFHQ+FFHQR scaling experiment (main paper and Sec.~\ref{sec:supp_ablation_tables}) shows the 600-image model is competitive with the full-data model on most metrics, the per-size sweep (Sec.~\ref{sec:supp_scale}) shows saturation rather than degradation around 600 images, and the three-seed robustness study (Tab.~\ref{tab:subset_robust}) shows the result is stable across random subsets.

\noindent\textbf{How sensitive is the method to degradation severity?} The Stage-2 fine-tuning uses the full Real-ESRGAN-style degradation range (Sec.~\ref{sec:supp_repro}); within that range the adapter robustly maps LQ latents into the HQ support. The failure cases (Sec.~\ref{sec:supp_quals}) are inputs that violate the aligned-frontal-face assumption (extreme occlusion, large yaw) or fall far outside the modelled degradation (JPEG $q{<}10$ with heavy blur), not merely strong-but-in-distribution degradations.

\noindent\textbf{Does the adapter add latency?} Negligibly. As the inference-cost breakdown in Sec.~\ref{sec:supp_repro} notes, the adapter is tiny relative to the single UNet step and the VAE codec, so one-step inference stays on par with the OSEDiff baseline.

\noindent\textbf{Could the adapter handle other inverse problems (impulse noise, box inpainting)?} In principle yes: the adapter never observes the degradation operator, only the latent the frozen VAE produces, so nothing in its formulation is specific to blur-plus-downsample-plus-JPEG. What would change is the size of the gap it must close (masking or impulse noise displaces the latent further than the degradations we train on) and therefore the codebook size and Stage-1 budget implied by Sec.~\ref{sec:supp_proof}. We have not run these settings and do not claim them.

\noindent\textbf{Why a quantised transport rather than a GAN?} An adversarial critic is a standard way to move one distribution onto another and would be a reasonable alternative to a codebook. We chose quantisation because at this parameter budget (5.2\,M, trained on 600 images for 100 epochs) an adversarial objective is the less stable of the two and would reintroduce the multi-stage fragility that motivated decoupling the adapter in the first place. A hybrid (codebook routing with an adversarial refinement on the mapped latent) is a plausible extension, and Proposition~1 does not argue against it: it argues only against \emph{global} corrections.

\noindent\textbf{Would an attribute-specific subset work as well as a random one?} Our 600 images are drawn uniformly at random, and the three-seed study shows the result does not depend on the particular draw. It does not show that a \emph{biased} draw would work: a subset restricted to one age group, skin tone, or hairstyle would cover fewer appearance modes, which by Proposition~1 is exactly the condition under which a fixed-$K$ codebook starts to pay a between-mode penalty on the modes it never saw. We regard the demographic composition of the training subset as an open and fairness-relevant question rather than a settled one.

\noindent\textbf{What are the main limitations?} LAFR assumes the strong geometric regularity of aligned faces; it should not be extrapolated to open-domain restoration, it is single-image (no temporal consistency for video), and rare degradations such as extreme occlusion may require additional degradation modeling. These are stated in the main paper and illustrated in Sec.~\ref{sec:supp_quals}.

\section{Societal Impacts}
\label{sec:supp_impact}

Our efficient face image restoration method can support beneficial applications: in medical diagnosis and forensic investigations, it can recover faces from partial or degraded data to assist identity verification and facial analysis in critical scenarios; for accessibility, it can improve facial expression recognition and avatar reconstruction quality, enhancing communication in virtual environments for individuals with disabilities; and for cultural heritage, it can help restore historical photographs.

However, the same capability may be misused for unauthorised reconstruction of individuals' faces, raising concerns about privacy invasion, surveillance, and identity fraud if not properly regulated. We encourage practitioners to deploy LAFR only within frameworks that include explicit consent, traceable provenance for restored images, and access controls. Watermarking the restored outputs and providing audit logs are practical safeguards that integrate cleanly with the lightweight nature of our pipeline.

\section{LLM Usage Declaration}
\label{sec:supp_llm}
In preparing this document we used a Large Language Model only for writing assistance: text polishing, grammar correction, and improving clarity. It was not used to generate research ideas, design experiments, or produce results. All content was reviewed, revised, and approved by the authors, who take full responsibility for its accuracy and originality.

\bibliography{main}
\end{document}